\documentclass[10pt,twocolumn,letterpaper]{article}
\usepackage[accsupp]{axessibility} 
\usepackage[]{cvpr}              
\usepackage[dvipsnames]{xcolor}
\definecolor{cvprblue}{rgb}{0.21,0.49,0.74}
\usepackage[pagebackref,breaklinks,colorlinks,citecolor=cvprblue]{hyperref}

\usepackage{graphicx}
\usepackage{amsmath}
\usepackage{amssymb}
\usepackage{booktabs}
\usepackage{CJKutf8}  
\usepackage{array}
\usepackage{tabularx}
\usepackage{multirow}
\usepackage{subcaption}
\usepackage{comment}
\usepackage{url}
\usepackage{colortbl}
\usepackage{nccmath}
\usepackage{makecell}

\usepackage{times}
\usepackage{epsfig}

\DeclareMathAlphabet\mathbfcal{OMS}{cmsy}{b}{n}

\usepackage[capitalize]{cleveref}
\crefname{section}{Sec.}{Secs.}
\Crefname{section}{Section}{Sections}
\Crefname{table}{Table}{Tables}
\crefname{table}{Tab.}{Tabs.}

\newcolumntype{Y}{>{\centering\arraybackslash}X}

\begin{document}
\begin{CJK}{UTF8}{}
\CJKfamily{mj}

\title{Differentiable Point-based Inverse Rendering}

\author{
Hoon-Gyu Chung ~ ~ ~
Seokjun Choi ~ ~ ~
Seung-Hwan Baek \\
POSTECH\\
}

\twocolumn[{%
\renewcommand\twocolumn[1][]{#1}%
\maketitle
}]

\begin{abstract}
We present differentiable point-based inverse rendering, DPIR, an analysis-by-synthesis method that processes images captured under diverse illuminations to estimate shape and spatially-varying BRDF. To this end, we adopt point-based rendering, eliminating the need for multiple samplings per ray, typical of volumetric rendering, thus significantly enhancing the speed of inverse rendering. 
To realize this idea, we devise a hybrid point-volumetric representation for geometry and a regularized basis-BRDF representation for reflectance.
The hybrid geometric representation enables fast rendering through point-based splatting while retaining the geometric details and stability inherent to SDF-based representations. The regularized basis-BRDF mitigates the ill-posedness of inverse rendering stemming from limited light-view angular samples. We also propose an efficient shadow detection method using point-based shadow map rendering.
Our extensive evaluations demonstrate that DPIR outperforms prior works in terms of reconstruction accuracy, computational efficiency, and memory footprint. 
Furthermore, our explicit point-based representation and rendering enables intuitive geometry and reflectance editing. 
\end{abstract}

\vspace{-5mm}
\section{Introduction}
\label{sec:intro}
Inverse rendering aims to estimate geometry and reflectance of real-world objects from a set of images, with applications including relighting, augmented and virtual reality, and object digitization. While it has been a longstanding challenge in computer vision and graphics, inverse rendering is still notoriously difficult.
A common approach to solve inverse rendering is to take the analysis-by-synthesis principle. That is, synthesized images using forward rendering are compared to input captured images, and through the iterative minimization of the difference between these images, geometry and reflectance are updated. Hence, the forward rendering plays a critical role in the analysis-by-synthesis inverse rendering framework.

Mesh-based rendering methods combined with path tracing are prevalent for photo-realistic forward rendering.
However, they often struggle in inverse rendering due to discontinuity issues~\cite{loubet2019reparameterizing}. Volumetric rendering has recently been recognized for view synthesis, particularly in variants of NeRFs~\cite{nerf,refnerf,neus}. 
Inspired by such success, many state-of-the-art inverse rendering methods have adopted volumetric rendering as a forward-rendering engine, demonstrating compelling reconstruction of geometry and reflectance~\cite{tensoir,nerv, psnerf,iron,nerfactor}. However, challenges remain, such as the excessive computational costs due to multiple sampling for each ray, and ambiguities in BRDF and normal reconstructions given limited light-view angular samples.

Point-based rendering has been extensively studied for forward rendering,  where points serve as compact scene representations~\cite{3dgaussian,kopanas2021point,adop,synsin}. Point-based forward rendering involves splatting these points onto a viewpoint using circular disks, ellipsoids, or surfels~\cite{surfels,pointshop,surfacesplatting, perspective}. 
Point-based rendering, especially in the analysis-by-synthesis framework, have recently gained renewed interest~\cite{kopanas2021point,pulsar, adop,synsin, differentiable, papr}: compared to volumetric rendering, point-based rendering is more efficient since it avoids the need for multiple samplings per ray. Recent methods proposed by Kopanas et al.\cite{catacaustics,kopanas2021point}, Point-NeRF\cite{pointnerf}, Zhang et al.\cite{differentiable}, and Kerbl et al.\cite{3dgaussian} have shown the effectiveness of point-based rendering in modeling radiance fields for novel-view synthesis.

Inspired by this success, we present differentiable point-based inverse rendering, DPIR, that exploits point-based forward rendering for inverse rendering.  
DPIR processes either multi-view multi-light images or photometric images captured by multi-view photometric setups~\cite{diligentmv} and flash photography, respectively. 
Using point-based forward rendering confronts several challenges critical for inverse rendering:
(1) The discrete point representation hinders the reconstruction of both smooth and detailed surface.
(2) The inherent difficulty in jointly reconstructing geometry and spatially-varying BRDF from limited light-view samples remains.
(3) There is a need for efficient shadow consideration to ensure precise inverse rendering.

To address these challenges, we develop a hybrid point-volumetric geometry representation and a regularized basis-BRDF representation. The hybrid geometric representation enjoys the benefits of both point-based and volumetric geometry representations, ensuring efficient rendering through point-based splatting, while benefiting from the intricate and stable modeling capability inherent to signed-distance functions (SDFs). Note that points are our only rendering primitives. The regularized basis-BRDF representation consists of a per-position diffuse albedo and a specular reflectance component.
The specular component is modeled with weighted specular-basis BRDFs by exploiting the spatial coherency of specular reflectance.
This allows for overcoming the ill-posedness of inverse rendering under limited light-view angular samples. 
Also, we tackle the shadow detection problem through an efficient point-based shadow detection method, bypassing the volumetric integration often used in learning-based inverse rendering methods~\cite{nerv,psnerf,nerfactor,invrender}. 
DPIR then jointly optimizes point locations, point radii, diffuse albedo, specular basis BRDFs, specular coefficients, and SDF, by leveraging point-based splatting as a forward renderer in the analysis-by-synthesis framework. 
Our extensive evaluations show that DPIR outperforms previous state-of-the-art inverse rendering methods~\cite{tensoir,psnerf,iron,physg}, in accuracy, training speed, and memory footprint.
Furthermore, the explicit point representation and rendering of DPIR enable convenient scene editing.

In summary, our contributions are:
\begin{itemize}
    \item DPIR, an inverse rendering method that exploits differentiable point-based rendering as a forward renderer, outperforming prior methods in reconstruction accuracy, training speed, and memory footprint. 
    \item Hybrid point-volumetric geometry representation and a regularized basis-BRDF representation that enable efficient and high-quality optimization of geometry and spatially-varying BRDFs through point-based rendering. 
    \item Point-based shadow detection method merged into our inverse rendering framework, enabling efficient and accurate shadow detection for inverse rendering. 
\end{itemize}
\begin{figure*}[!ht]
	\centering
		\includegraphics[width=\linewidth]{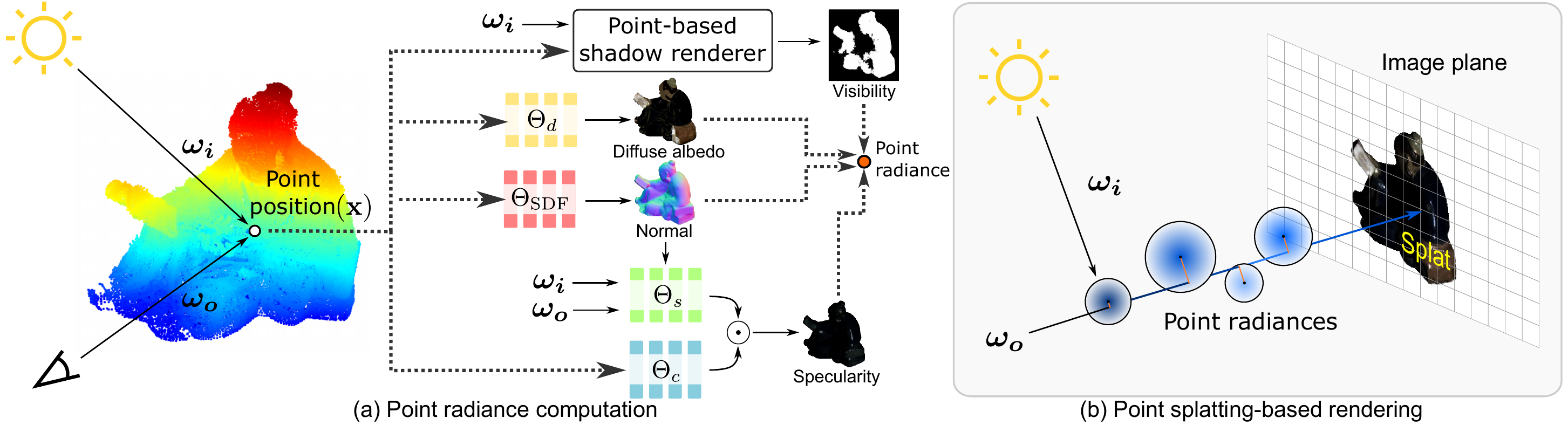}
		\caption{\textbf{Overview of differentiable forward rendering.} (a) For each 3D point, its position is used as a query for the diffuse-albedo MLP $\Theta_d$, SDF MLP $\Theta_\text{SDF}$, and specular-basis coefficient MLP $\Theta_c$. The specular-basis BRDF MLP $\Theta_s$ models specular-basis reflectance, given the incident and outgoing directions $\boldsymbol{\omega_{i}}$ and $\boldsymbol{\omega_{o}}$. The point-based shadow renderer estimates the point visibility from a light source per each image. By using the diffuse albedo, normals, specular reflectance, and visibility, we compute the radiance for each point. (b) The radiance is then projected onto a camera plane to render the pixel color through splatting-based differentiable forward rendering.}
  		\label{fig:overview}
    \vspace{-5mm}
\end{figure*}

\section{Related Work}
\label{sec:related}
\paragraph{Inverse Rendering}
Inverse rendering is a long-standing problem in computer vision and graphics~\cite{barron2014shape,dong2014appearance,shape,mvps,efficient,practical,schmitt2020joint,xia2016recovering,mvpsiso}.
Learning-based single-image inverse rendering enables accurate reconstruction for planar samples~\cite{chen2021dib,indoorir,li2018learning,homeir, sang2020single,sengupta2019neural,wei2020object, yu2019inverserendernet}.
For non-planar objects, Li et al.\cite{shadow} proposed an implicit neural representation that estimates geometry and reflectance from single-view multi-light images, albeit suffering from depth ambiguity.
For multi-view inputs captured under a constant lighting, learning-based inverse rendering methods~\cite{nerv,nerfactor} have recently shown promising results by exploiting volumetric rendering as in NeRF~\cite{nerf}.
Incorporating a signed distance function (SDF) as a geometric representation into volumetric rendering further improves normal-reconstruction quality~\cite{nero,iron,physg,invrender}.
However, these methods struggle with low reconstruction accuracy of spatially-varying reflectance and excessive training time due to the multiple sampling per each ray in the volumetric rendering process.
Monte Carlo differentiable rendering can be also used for inverse rendering, however it suffers from complex scene geometry~\cite{cai2022physics,hasselgren2022shape,luan2021unified} and limited light-view angular samples~\cite{hasselgren2022shape,munkberg2022extracting,sun2023neural}, due to Monte Carlo noise and mesh representations.

Using multi-view and multi-light inputs improves reconstruction quality as demonstrated in PS-NeRF~\cite{psnerf}. 
However, volumetric rendering used in PS-NeRF still results in excessive training time, which is further complicated by its multi-stage pipeline that requires a pretrained photometric stereo network. MVPSNet~\cite{mvpsnet} demonstrated faster training time by extracting lighting features for multi-view photometric stereo. 
However, only geometry is estimated that cannot be used for inverse rendering applications where reflectance is necessary.

Our DPIR provides highest reconstruction quality, fastest training time, and lowest memory footprint.
To this end, we use points as rendering primitives, resulting in fast training with splatting-based rendering. To handle the limited light-view angular samples, we gather multiple samples from points with similar specular appearance by using regularized basis BRDFs. Also, we handle shadow using efficient point-based shadow rendering. 

\paragraph{Point-based Rendering}
Point-based rendering uses points as a compact scene representation for rendering~\cite{pointsample}.
Splatting with circular disks, ellipsoids, or surfels~\cite{surfels,pointshop, surfacesplatting,perspective} enables high-quality and efficient point-based rendering.
To model geometry and outgoing radiance distribution of a scene, various point-based methods have been proposed.
Combining point-based rendering with image-refinement neural networks enables high-quality novel view synthesis~\cite{npbg,catacaustics,kopanas2021point,pulsar,lightfield, npbg++,adop, synsin,viewpoints}. 
Recently, Kopanas et al.\cite{catacaustics} achieve high-quality novel-view synthesis of highly specular surfaces using point-based rendering and explicitly modeling reflection.
Zhang et al.\cite{differentiable} employ efficient point-based rendering and spherical-harmonics point radiance for efficient novel view synthesis, without using refinement neural networks.
Kerbl et al.\cite{3dgaussian} use anisotropic Gaussians and tile-based optimization method for efficient and accurate novel-view synthesis of in-the-wild scenes. 
Yifan et al.\cite{geometry} employ geometric regularizers that lead to high-quality geometry reconstruction with point-based rendering.

In summary, point-based rendering has proven itself as a promising technique that has fast rendering speed, low memory footprint, and high rendering quality. 
In this paper, we combine the point-based rendering with the analysis-by-synthesis inverse rendering framework. 
By resolving associated challenges, DPIR demonstrates high-quality and efficient inverse rendering. 

\section{Method}
\label{sec:method}

The proposed DPIR estimates points, surface normals, spatially-varying BRDF, and visibility from the images of a static object captured either by flash photography~\cite{iron} or multi-view multi-light imaging~\cite{psnerf}. We use object masks as additional inputs similar to NeuS~\cite{neus} and Zhang et al.~\cite{differentiable}.
DPIR compares rendered images with the input images to update our scene representation. Figure~\ref{fig:overview} shows a detailed overview of our differentiable forward rendering process, which is pivotal for achieving efficient and high-quality inverse rendering.

\subsection{Scene Representation}

\paragraph{Hybrid Point-volumetric Geometric Representation}
We use a set of 3D points, where each point possesses two parameters: position $\mathbf{x} \in \mathbb{R}^{3\times1}$ and radius ${r}\in \mathbb{R}^{1\times1}$. 
Using the points as geometric primitives allows for fast splatting-based rendering by bypassing the per-ray integration used in volumetric rendering. 
However, using the points only often makes its surface normals noisy as the discrete points are non-uniformly distributed through space.
To represent both detailed and smooth geometry with accurate surface normals $\mathbf{n}$ for each point at position $\mathbf{x}$, we use SDF, represented as a coordinate-based MLP, ${\Theta}_\mathrm{SDF}$, as follows:
\begin{equation}\label{eq:normal}
\mathbf{n} =\boldsymbol{\triangledown_{\mathbf{x}}}{\Theta}_\mathrm{SDF}\!\left (\mathbf{x} \right ),
\end{equation}
where ${\triangledown_{\mathbf{x}}}$ is the differentiation operator~\cite{neus,idr}. 

Our hybrid point-volumetric representation allows \emph{discrete points} to move in space while the surface normals of the points can be sampled from the \emph{continuous SDF}.

\paragraph{Regularized Basis BRDF Representation}
Reconstructing per-point BRDF from limited light-view angular samples is an ill-posed problem.
Thus, we propose to exploit spatial coherency of specular reflectance by using the basis BRDF representation~\cite{lawrence2006inverse,shadow,practical}.
Specifically, we use three MLPs for diffuse albedo, specular coefficients, and specular basis BRDFs, describing the total reflectance $f_r$ as 
\begin{equation}\label{BRDF}
f_{r}\! \left ( \boldsymbol{\omega_{o}}, \boldsymbol{\omega_{i}}, \mathbf{x}, \mathbf{n} \right ) = {\Theta}_{d}(\mathbf{x}) + {\Theta}_{s}\! \left ( \boldsymbol{h},\boldsymbol{n} \right ) {\Theta}_{c}(\mathbf{x}) ,
\end{equation}
where ${\Theta}_{d}$, ${\Theta}_{s}$, and ${\Theta}_{c}$ are the MLPs for the diffuse albedo, specular basis, and regularized specular-basis coefficients, respectively.
The outputs of the MLPs are in the following dimensions: ${\Theta}_{d}(\mathbf{x}) \in \mathbb{R}^{3\times1}$, ${\Theta}_{s}(\boldsymbol{h},\boldsymbol{n}) \in \mathbb{R}^{3\times K}$, ${\Theta}_{c}(\mathbf{x}) \in \mathbb{R}^{K\times1}$.
$K$ is the number of basis BRDFs. 
$\boldsymbol{\omega_{i}}$ and $\boldsymbol{\omega_{o}}$ are the incident and outgoing light directions.
$\mathbf{h}$ is the half-way vector: $\boldsymbol{h}=({\boldsymbol{\omega_{i}}+\boldsymbol{\omega_{o}}})/{\left\|\boldsymbol{\omega_{i}}+\boldsymbol{\omega_{o}}\right\|}$.

\begin{figure}[t]
    \centering
    \includegraphics[width=0.85\linewidth]{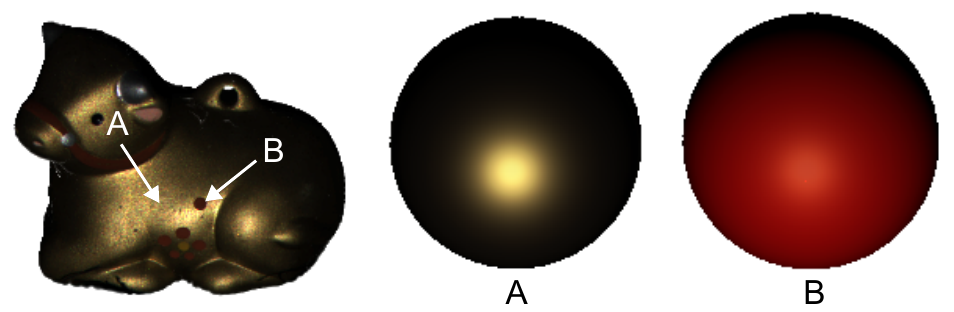}
    \caption{\textbf{Estimated spatially-varying BRDFs.} DPIR accurately reconstruct the BRDFs of the specular gold appearance and the red diffuse appearance. We visualize the BRDFs on unit spheres illuminated by a point light source. }
    \label{fig:basis}
    \vspace{-5mm}
\end{figure}
Unlike previous inverse rendering methods utilizing basis BRDFs~\cite{shadow,practical,psnerf}, we found that enforcing a positive constraint in the \emph{specular basis coefficients} and optimizing under an $\boldsymbol{l_{1}}$-norm with a lower bound of $\epsilon$ significantly enhances the accuracy of reflectance estimation. {We set $\epsilon=1$ for highly glossy objects, $\epsilon=0.5$ otherwise.}
Figure~\ref{fig:basis} shows the learned BRDFs consisting of per-point diffuse albedo and regularized specular basis BRDFs. 

\subsection{Point-based Visibility Test}
\begin{figure}[t]
    \centering
    \includegraphics[width=\linewidth]{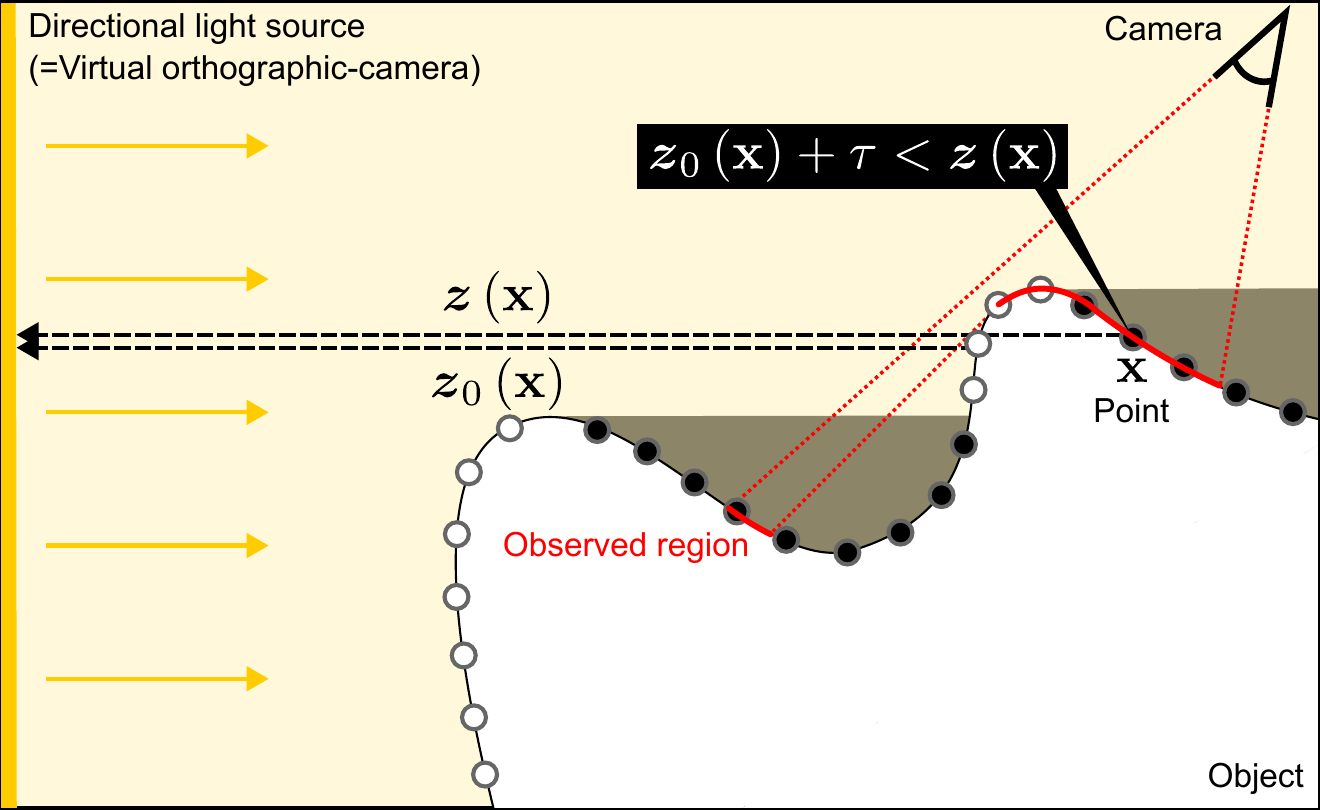}
    \caption{\textbf{Point-based visibility test.} To determine the visibility of each point, we compute the depth using the z-buffer from a virtual camera positioned at the light source, and compare the depth with the distance from each point to the virtual camera.
}
\vspace{-4mm}
\label{fig:shadow_overview}
\end{figure}

Identifying shadow on which illumination does not reach to a point is critical for robust inverse rendering~\cite{tensoir,nerv,psnerf,nerfactor,invrender}.
Departing from computationally extensive volumetric-rendering approaches for visibility test in recent learning-based methods~\cite{shadow,psnerf,nerfactor}, we use a simple method compatible with our point representation: a shadow map technique used in rasterization-based graphics~\cite{hourcade1985algorithms,reeves1987rendering, shadowcasting}.
Figure~\ref{fig:shadow_overview} shows the schematics of our visibility test.
For each input image and a known light source, we place a virtual orthographic camera at the location of the directional light source. 
We then perform splatting-based rendering of the points to the virtual camera.
The resulting z-buffer stores the depth $\boldsymbol{z}$ of each point with respect to the virtual camera.
The visibility function $f_{v}$ then can be estimated from the z-buffer as 
\begin{equation}
\label{eq:shadow}
f_{v}\left (\boldsymbol{\omega_{i}},\mathbf{x} \right) = \sigma\left ( \tau + \boldsymbol{z}_{0}\left(\mathbf{x}\right)-\boldsymbol{z}\left (\mathbf{x}\right ) \right), 
\end{equation}
where $\sigma\left (\cdot \right )$ is the step function that returns 1, meaning visible from the light source if the input is positive, and 0, which means invisible, otherwise. 
$\boldsymbol{z}_{0}\left(\mathbf{x}\right)$ denotes the depth value of the first intersection point and $\tau$ is the threshold, set to be 0.1 in our experiments. 
{For photometric dataset, visibility of every point is set to 1 as the light source and the camera are co-located.}

Our shadow handling method efficiently and accurately compute visibility of each point with point-based rasterization, which brings benefits over existing learning-based shadow estimation that requires computationally expensive volumetric ray marching~\cite{shadow}. 

\subsection{Optimization via Point Splatting}
\paragraph{Point-based Rendering}
For rendering each image, we use the point location $\mathbf{x}$, normals $\mathbf{n}$ sampled from the SDF, reflectance from the learned BRDFs $f_r(\cdot)$, and visibility from the point-based test $f_v(\cdot)$. 
We compute the radiance $R$ for each point $\mathbf{x}$ along the direction $\boldsymbol{\omega_o}$ towards a camera pixel as follows:
\begin{equation}
\label{eq:simplerendering}
 R\!\left ( \boldsymbol{\omega_{o}},\mathbf{x}\right)\!=\!L\!\left(\boldsymbol{\omega_{i}},\mathbf{x}\right)\! f_{v}\!\left 
(\boldsymbol{\omega_{i}},\mathbf{x} \right)\! f_{r}\! \left ( \boldsymbol{\omega_{o}},
\boldsymbol{\omega_{i}},\mathbf{x}, \mathbf{n} \right )\!\left (\boldsymbol{\omega_{i}}\cdot \mathbf{n} \right),
\end{equation}
where $L(\boldsymbol{\omega_{i}},\mathbf{x})$ is the incident radiance from the direction $\boldsymbol{\omega_{i}}$ to the point $\mathbf{x}$.
Equation~\eqref{eq:simplerendering} is a simplified rendering equation by exploiting that a scene point in each input image is contributed from a single incident light ray, which is valid for both multi-view multi-light images~\cite{diligentmv} and photometric images.
Note that extending the rendering equation to handle multiple incident light rays is feasible by simply integrating the computed radiance, which we demonstrate in the environment-map rendering results.

Once the radiance of each point is computed, we project the points to the camera viewpoints from which images are captured, using point splatting.
The rendered pixel intensity of a camera pixel $u$ amounts to the results of $\alpha$-blending of the projected point radiance $R$:
\begin{align}\label{eq:pixelcolor}
{I}\left ( u \right )&=\sum_{i=1}^{N}  R_i \alpha_{i}(u)\prod_{j=1}^{i-1}
\left( 1-\alpha_{j}\left ( u \right )\right ), \\
\alpha_{i}\left ( u \right ) &= 1 - \frac{\left (p_{i}-u \right)^{2}}{r_i^{2}},
\end{align}
where $N$ is the number of points, $R_i$ is the radiance of $i$-th point computed from Equation~\eqref{eq:simplerendering}, $p_i$ is the projected pixel location, and $r_i$ is the radius of the point.
$\alpha_i$ is the corresponding weight. 
{Points are sorted according to their distances to the camera}.
This point-based rendering equation, also depicted in Figure~\ref{fig:overview}, can be efficiently evaluated, making DPIR outperform previous volumetric inverse rendering methods in efficiency.

\paragraph{Optimization}
We propose to optimize point positions $\mathbf{x}$, point radii $r$, and MLPs for SDF $\Theta_\text{SDF}$, diffuse albedo $\Theta_d$,  specular coefficients $\Theta_c$, and specular-basis BRDFs $\Theta_s$ by minimizing the following loss:
\begin{align}\label{eq:loss}
\mathcal{L}_2 + \lambda_\mathrm{ssim} \mathcal{L}_\mathrm{ssim}
+\lambda_{\mathrm{SDF}}  \mathcal{L}_\mathrm{SDF} + \lambda_{\mathrm{c}}   \mathcal{L}_\mathrm{c} +\lambda_{\mathrm{m}} \mathcal{L}_\mathrm{m},
\end{align}
where $\mathcal{L}_2$ and $\mathcal{L}_\mathrm{ssim}$ are the $\boldsymbol{l_{2}}$ loss and the SSIM loss for the rendered image $I$ and the captured image $I'$.

$\mathcal{L}_\mathrm{SDF}$ promotes the zero-level set of the SDF lies near the point positions: $\mathcal{L}_\mathrm{SDF} = \left\|{\Theta}_\mathrm{SDF}\!\left (\mathbf{x} \right)\right\|^{2}.$
$\mathcal{L}_\mathrm{c}$ regularizes  $\boldsymbol{l_{1}}$-norm of per point specular coefficient to be $\epsilon$:  $\mathcal{L}_\mathrm{c} = \left\| \Theta_{c}(\mathbf{x})\right\|_{1}-\varepsilon.$ 
{$\mathcal{L}_\mathrm{m}$ is defined as the $\boldsymbol{l_{2}}$ loss between the estimated mask and the ground-truth mask, aligning the rendered mask from the points to the input masks.} 
For stable optimization, we employ the mask-based initialization of the point locations, pruning, and upsampling schemes following Zhang et al.~\cite{differentiable}. 
We refer to the Supplemental Document for the details of the loss functions and optimization techniques. 
\begin{figure*}[t]
    \centering
    \includegraphics[width=\linewidth]{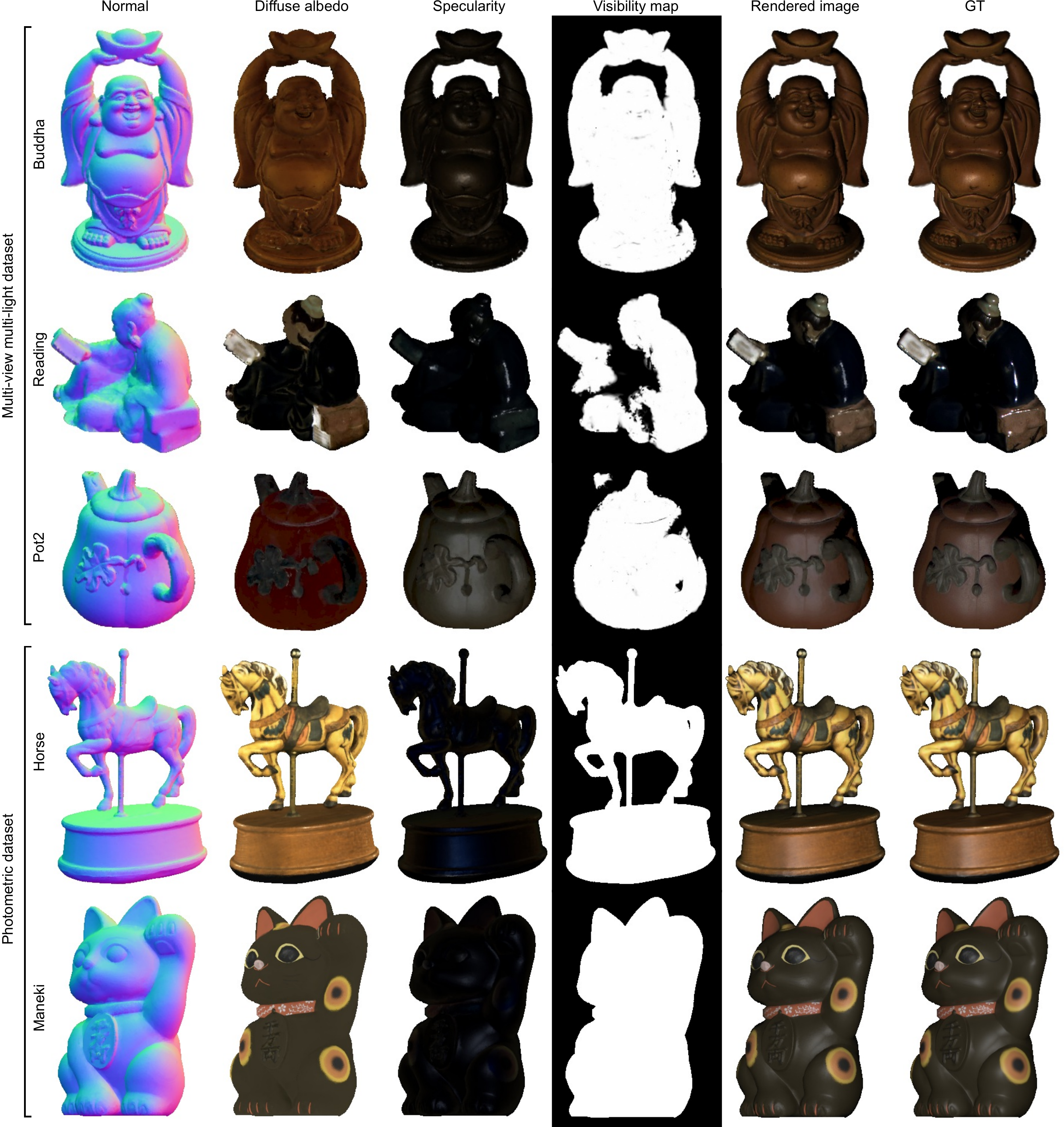}
    \caption{\textbf{Reconstruction results.} DPIR enables accurate and efficient reconstruction of geometry and reflectance for various objects.}
\vspace{-3mm}
\label{fig:results} 
\end{figure*}

\section{Results}
\label{sec:Experiment}
DPIR enables accurate and efficient reconstruction of geometry and reflectance as demonstrated in Figure~\ref{fig:results}. In the following, we evaluate DPIR compared to previous methods and provide detailed ablation results.

\begin{table*}[t]
\resizebox{2.05\columnwidth}{!}{
    \begin{tabular}{c|c|ccc|ccc|ccc|ccc|ccc}
    \toprule[1pt]
            &\multicolumn{1}{c|}{}&\multicolumn{3}{c|}{Bear} & \multicolumn{3}{c|}{Buddha} & \multicolumn{3}{c|}{Cow} & \multicolumn{3}{c|}{Pot2} & \multicolumn{3}{c}{Reading} \\ \hline
    Config.&Method &PSNR $\uparrow$&SSIM $\uparrow$&LPIPS $\downarrow$           &PSNR$\uparrow$&SSIM $\uparrow$&LPIPS $\downarrow$              &PSNR $\uparrow$&SSIM $\uparrow$&LPIPS $\downarrow$           &PSNR $\uparrow$&SSIM $\uparrow$&LPIPS $\downarrow$            &PSNR $\uparrow$&SSIM $\uparrow$&LPIPS$\downarrow$  \\
    \hline
    &PhySG     & 24.52 & 0.9590 & 0.041 & 20.92 & 0.9229 & 0.0624  & 23.64 &  0.9604 & 0.0342  & 25.21 & 0.9609 & 0.0241 & 19.05   & 0.9056   & 0.0817  \\
    Single &TensoIR   & 24.81 & 0.9597 & 0.0510  & 25.40 & 0.9521 & 0.0374  & 27.09 & 0.9766 & 0.0260 & 26.89  & 0.9741  & 0.0309 & 27.02 & 0.9607 & 0.0277   \\ 
    light &PS-NeRF  & 35.19 & 0.9925 & 0.0061 & 32.85   & 0.9783   & 0.0074  & 36.57 & 0.9942 & 0.0026  & 38.40 & 0.9943 & \textbf{0.0019}  & 33.73   & 0.9792  & 0.0077   \\
    &Ours  & \textbf{43.21} & \textbf{0.9944} & \textbf{0.0037} & \textbf{37.62} & \textbf{0.9858} & \textbf{0.0048} & \textbf{38.23} & \textbf{0.9946} & \textbf{0.0022} & \textbf{39.32} & \textbf{0.9951} & 0.0022 & \textbf{35.75} & \textbf{0.9843} & \textbf{0.0063}    \\  \hline
    Multi &PS-NeRF  & 34.27 & 0.9802 & 0.0127& 31.58 & 0.9637 & 0.0114 & 36.03 & 0.9871 & 0.0066  & 37.76 & 0.9851 & 0.0041  & 31.16 & 0.9736 & \textbf{0.0202}    \\
    light &Ours  & \textbf{39.78} & \textbf{0.9821} & \textbf{0.0083} & \textbf{34.88} & \textbf{0.9726} & \textbf{0.0090} & \textbf{37.64} & \textbf{0.9890} & \textbf{0.0041} & \textbf{38.86} & \textbf{0.9885} & \textbf{0.0034} & \textbf{32.50} & \textbf{0.9788}  & 0.0204  \\ 
    \bottomrule[1pt]
    \end{tabular}}
    \caption{Quantitative comparison of novel-view rendering and relighting on DiLiGenT-MV dataset.
 }
\vspace{-4mm}
\label{tab:rendering_comparison}
\end{table*}

\begin{table}[h]
\resizebox{1.0\columnwidth}{!}{
    \begin{tabular}{c|ccccc|cc}
    \toprule[1pt]
           &\multicolumn{1}{c}{Bear} & \multicolumn{1}{c}{Buddha} & \multicolumn{1}{c}{Cow} & \multicolumn{1}{c}{Pot2} & \multicolumn{1}{c}{Reading} & \multicolumn{1}{c}{Train} &\multicolumn{1}{c}{Mem.} \\
    Method  & {MAE $\downarrow$}   & {MAE $\downarrow$}    & {MAE $\downarrow$}    & {MAE $\downarrow$}   & {MAE $\downarrow$}   &time$\downarrow$  &$\downarrow$   \\ 
    \hline
    PhySG                   & 11.35 & 27.20 & 16.10 & 11.98 & 26.67 & 20h  & 13MB \\
    TensoIR                 & 35.93 & 37.54 & 30.10 & 25.68 & 33.67 & 6h  & 68MB \\ 
    PS-NeRF                 & 4.68 & 11.92 & 5.89 & 7.55 & 10.64 & 22h  & 40MB     \\ 
    \hline
    Ours                   & \textbf{4.35} & \textbf{11.10} & \textbf{4.61} & \textbf{6.71} & \textbf{9.03} & \textbf{2h}  & \textbf{5MB}     \\ 
    \bottomrule[1pt]
    \end{tabular}}
    \caption{Quantitative comparison of normal accuracy, average training time, and memory footprint on DiLiGenT-MV dataset.
}
\vspace{-2mm}
\label{tab:normal_comparison}
\end{table}

\begin{figure}[t]
\centering
\includegraphics[width=\linewidth]{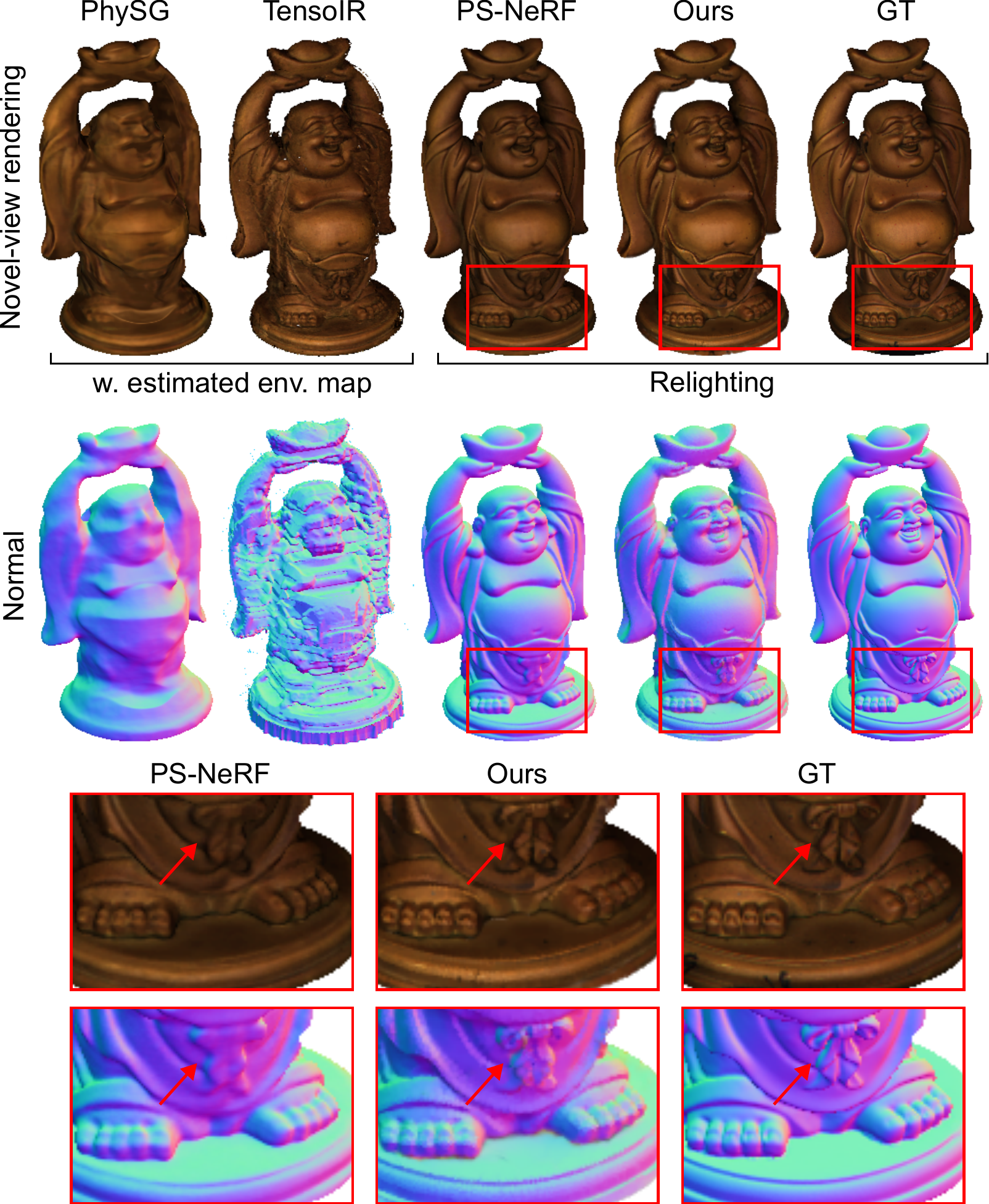}

\caption{\textbf{Comparison of novel view rendering and estimated normal on DiLiGenT-MV dataset.} Our DPIR recovers detailed surface normals and reproduces accurate appearance.}
\vspace{-4mm}
\label{fig:diligent_result}
\end{figure}

\subsection{Comparison}

\paragraph{Multi-view Multi-light Images}
We evaluate DPIR on DiLiGenT-MV~\cite{diligentmv}, a multi-view multi-light dataset, compared to state-of-the-art neural inverse rendering methods: PhySG~\cite{physg}, TensoIR~\cite{tensoir} and PS-NeRF~\cite{psnerf}. 
Table~\ref{tab:rendering_comparison} and Table~\ref{tab:normal_comparison} show quantitative evaluations of novel-view relighting, normal accuracy, training time, and memory footprint. 
DPIR not only outperforms the compared methods in rendering and normal accuracy, but also offers 10$\times$ faster training and 8$\times$ lower memory footprints than PS-NeRF, which is the only competitive method in rendering accuracy. 
Figure~\ref{fig:diligent_result} shows novel-view rendering images and estimated normals. 
Note that DPIR does not require using any pre-trained network contrary to PS-NeRF that uses a pre-trained photometric-stereo network.

For the above experiments, we have two comparison configurations.
First, since PhySG and TensoIR assume constant environment illumination, we use the multi-light averaged image for each view to simulate a virtual environment map~\cite{psnerf}. 
PhySG and TensoIR are trained on 15 views lighting-averaged images and tested on 5 novel views under the same environment map. 
As PS-NeRF and our DPIR take varying illumination images, we use 15 views and 16 lightings for training and use 5 views and 96 lightings for testing. 
For the comparison with PhySG and TensoIR, we compute the average image across 96 lightings for PS-NeRF and DPIR on each testing view.  
This comparison protocol is adopted from the PS-NeRF~\cite{psnerf}.
Second, we make a comparison only between PS-NeRF and DPIR to evaluate true novel-view relighting. 
That is, we directly compare the rendered images under a novel view and lighting without any averaging. 
Our DPIR achieves the best performance over the previous methods in both configurations.

\begin{figure}[t]
\centering
\includegraphics[width=\linewidth]{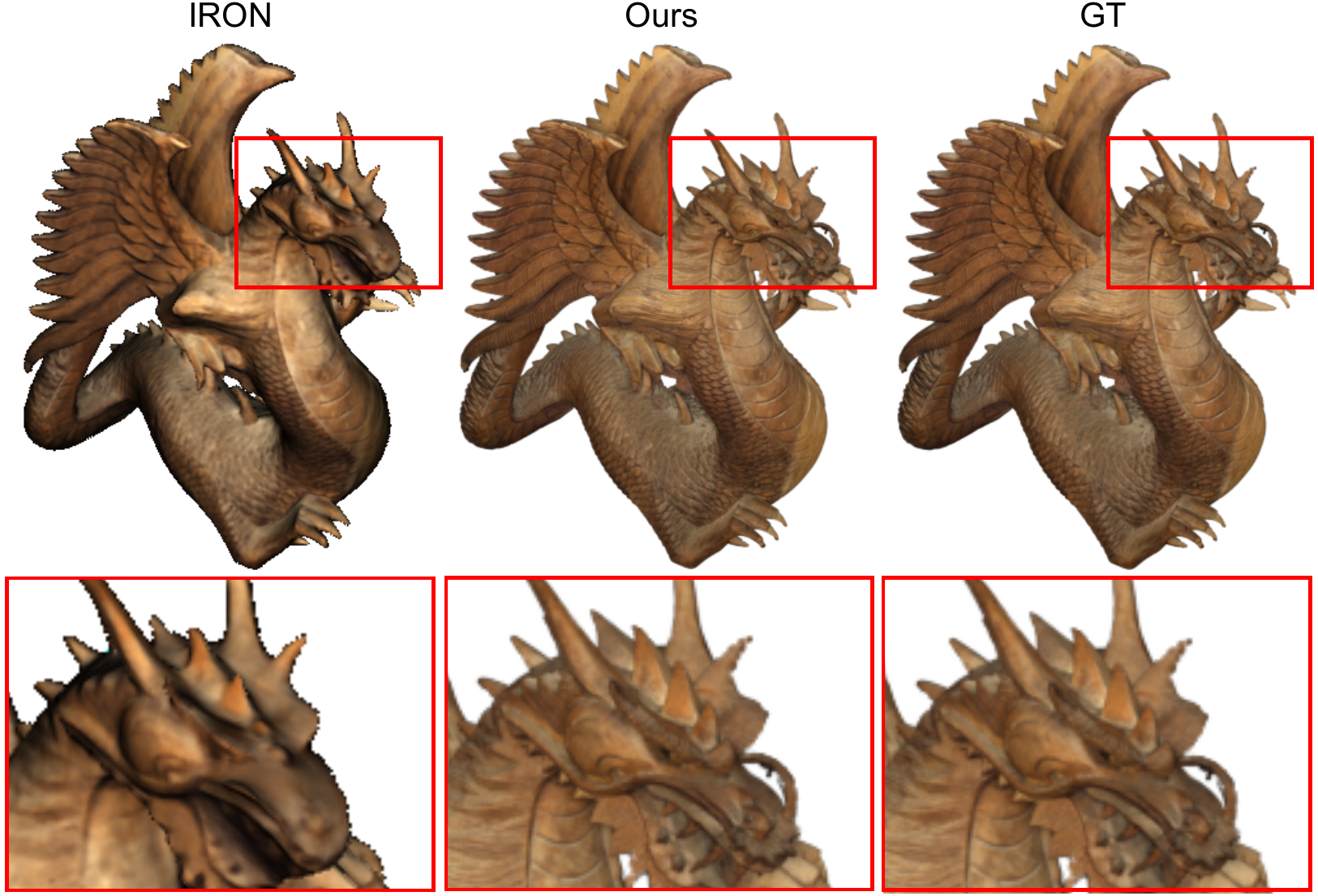}

\caption{\textbf{Comparison of novel-view flashlight relighting.} DPIR successfully reproduces the ground-truth appearance, outperforming IRON.}
\vspace{-3mm}
\label{fig:photo_result}
\end{figure}

\begin{table}[t]
    \centering
    \resizebox{0.9\columnwidth}{!}{
    \begin{tabular}{c|cccc|cc}
        \toprule[1pt]
             & PSNR$\uparrow$ & SSIM$\uparrow$ & LPIPS$\downarrow$ & MAE$\downarrow$ &  Train$\downarrow$ & Mem$\downarrow$   \\ \hline
        IRON & 31.91    & 0.9557     &   0.0446     &   9.38     &  12h   & 28MB   \\ \hline
        Ours &  \textbf{35.56} & \textbf{0.9734}  & \textbf{0.0285}   &   \textbf{8.74}   &  \textbf{2h}   & \textbf{5MB}   \\ 
        \bottomrule[1pt]  
        \end{tabular}}
        \caption{Quantitative comparison of novel-view relighting, normal accuracy, training time, and memory footprint on the photometric dataset.}
    \vspace{-4mm}
    \label{tab:photo_comparison}
\end{table}

\paragraph{Photometric Images}
We then evaluate DPIR on a photometric dataset, which we render using four objects at 300 views with co-located point lights, following the configuration of mobile flash photography~\cite{practical}. 
200/100 views are used for training/testing, respectively.
Table~\ref{tab:photo_comparison} and Figure~\ref{fig:photo_result} show that DPIR outperforms IRON~\cite{iron}, the-state-of-the-art inverse rendering method for photometric images, with higher novel-view relighting quality, 6$\times$ faster training, and 5$\times$ lower memory footprint.
IRON is a two-stage method that uses volumetric and surface renderings for each stage, resulting in not only lengthy training time, but also smoothed geometry and saturated reflectance. DPIR is based on one-stage point-based analysis-by-synthesis framework that directly optimizes geometry and reflectance, enabling fast training and accurate reconstruction.

\subsection{Ablation Study}

\begin{figure}[t]
    \centering
    \includegraphics[width=\linewidth]{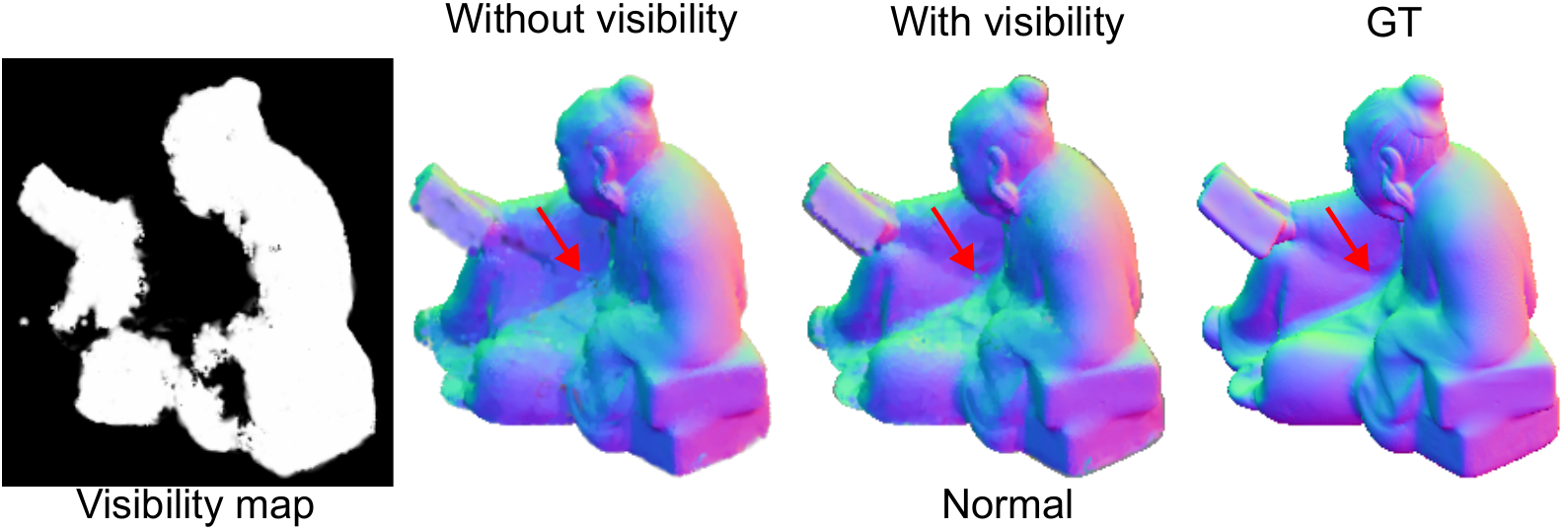}
    \caption{\textbf{Importance of point-based visibility test.} Our efficient point-based visibility test enables accurate normal reconstruction for occluded regions.
    }
\vspace{-0mm}
\label{fig:shadow} 
\end{figure}

\begin{figure}[t]
    \centering
    \includegraphics[width=\linewidth]{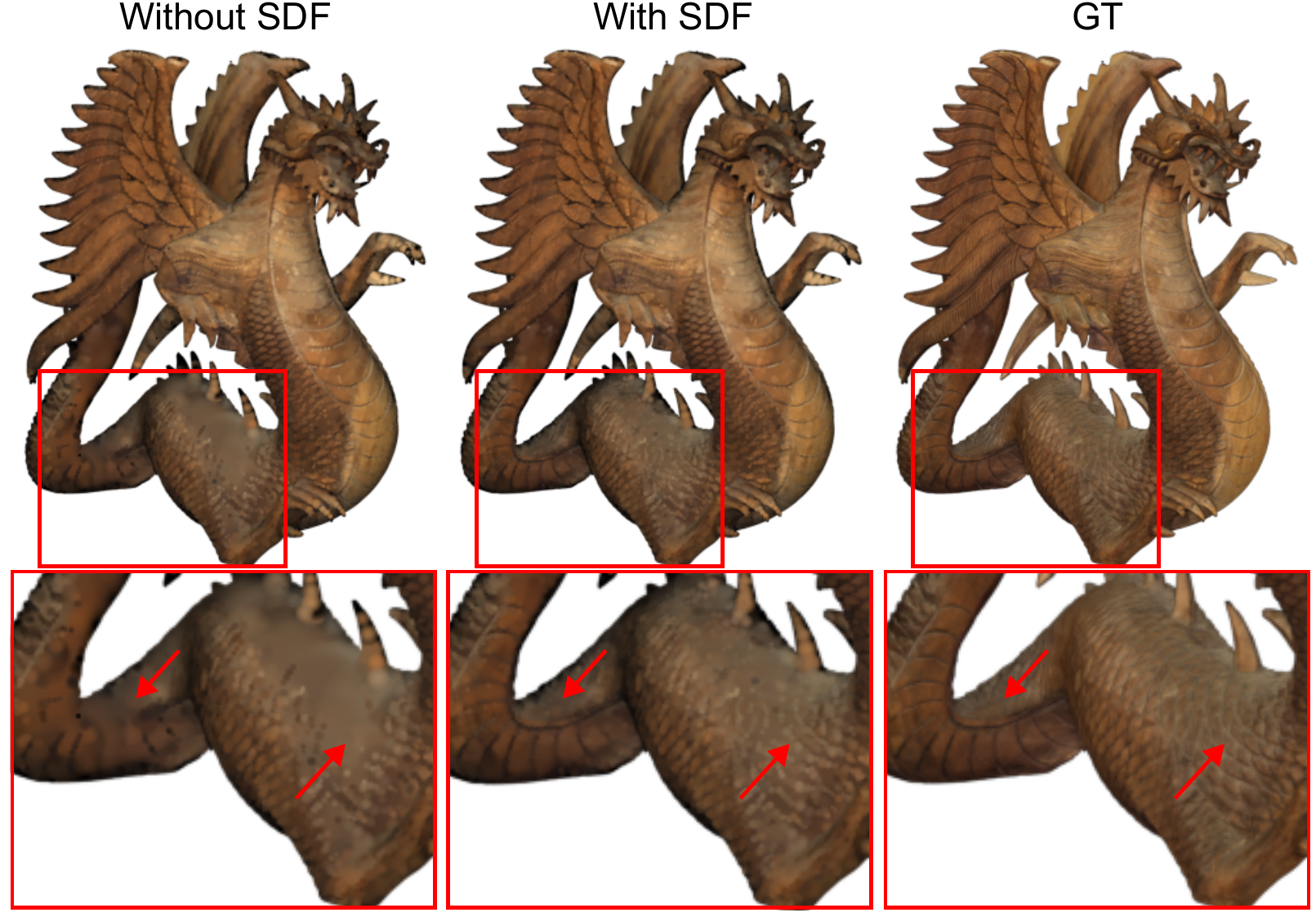}
    \caption{\textbf{Impact of hybrid shape representation.} Using points only as shape representation often results in inaccurate reconstruction. 
    Our hybrid shape representation based on points and a SDF allows for recovering detailed appearance. {We recommend viewing this figure on a monitor.}
    } 
\vspace{-4mm}
\label{fig:ablation_SDF} 
\end{figure}

\paragraph{Point-based Shadow Detection}
The point scene representation and rendering of DPIR allows for efficient and accurate shadow estimation, which is crucial for reconstructing objects with complex geometry. 
Our visibility test only requires 0.1$\times$ additional training time.
Figure~\ref{fig:shadow} shows that our point-based shadow detection leads to notable improvements of surface normal accuracy for the occluded region.

\paragraph{Hybrid Shape Representation}
DPIR utilizes the hybrid point-volumetric shape representation. Figure~\ref{fig:ablation_SDF} shows that only using point representation leads to unstable reconstruction. Our hybrid point-volumetric representation improves reconstruction quality by learning delicate geometry and appearance of both smooth and detailed surfaces. 

\begin{figure}[t]
    \centering
    \includegraphics[width=\linewidth]{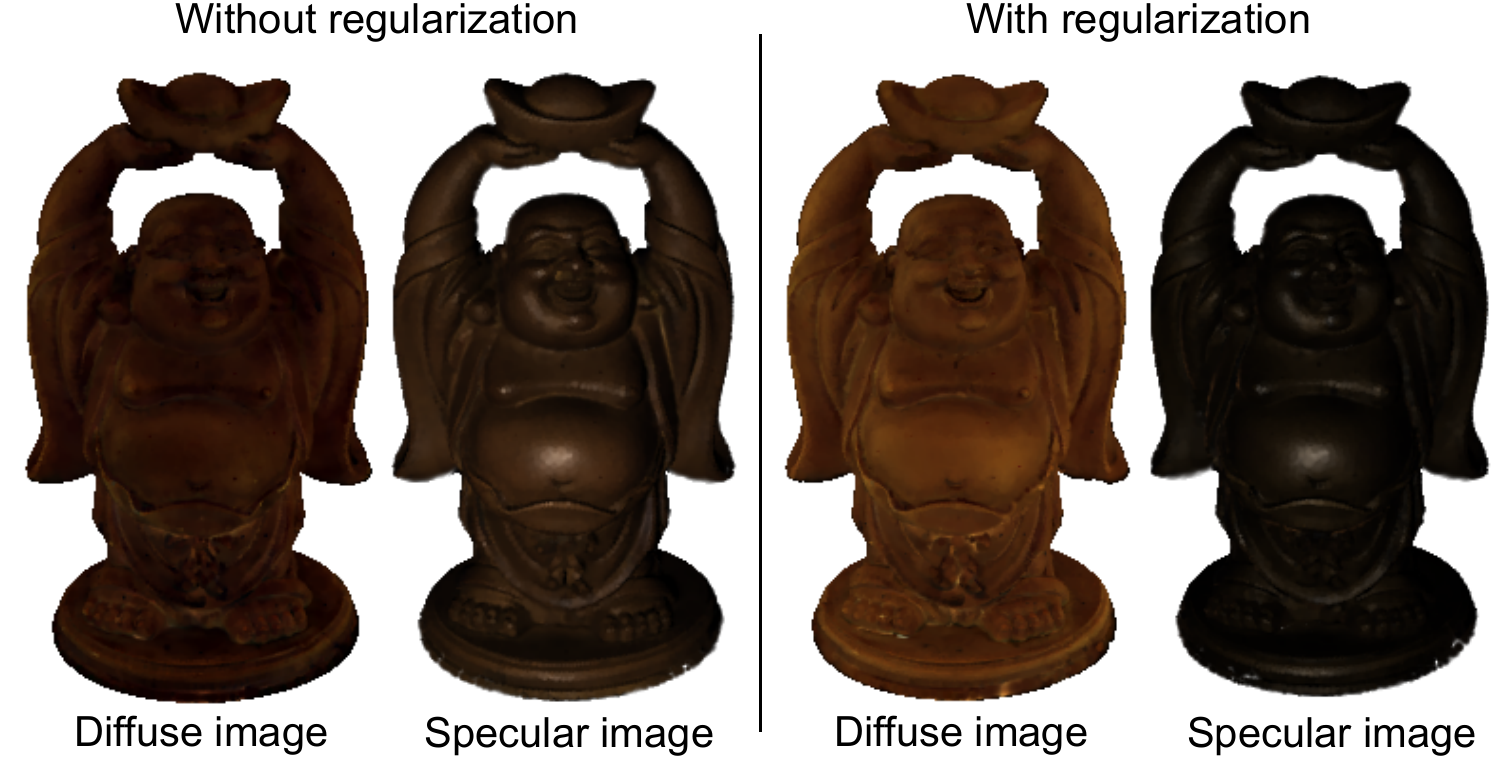}
    \caption{\textbf{Impact of basis-BRDF regularization.} Without using the regularization, diffuse appearance is baked into the specular image. Regularized specular coefficients enable accurate appearance decomposition. }
\vspace{-5mm}
\label{fig:specular_reg} 
\end{figure}

\paragraph{Regularization for Specular Basis Coefficients}
DPIR constrains specular coefficients with the regularization loss.
Figure~\ref{fig:specular_reg} shows that using the regularizer leads to accurate appearance decomposition between diffuse and specular components. 
Without using the regularizer, diffuse appearance is baked into the estimated specular image.

\begin{figure}[t]
    \centering
    \includegraphics[width=\linewidth]{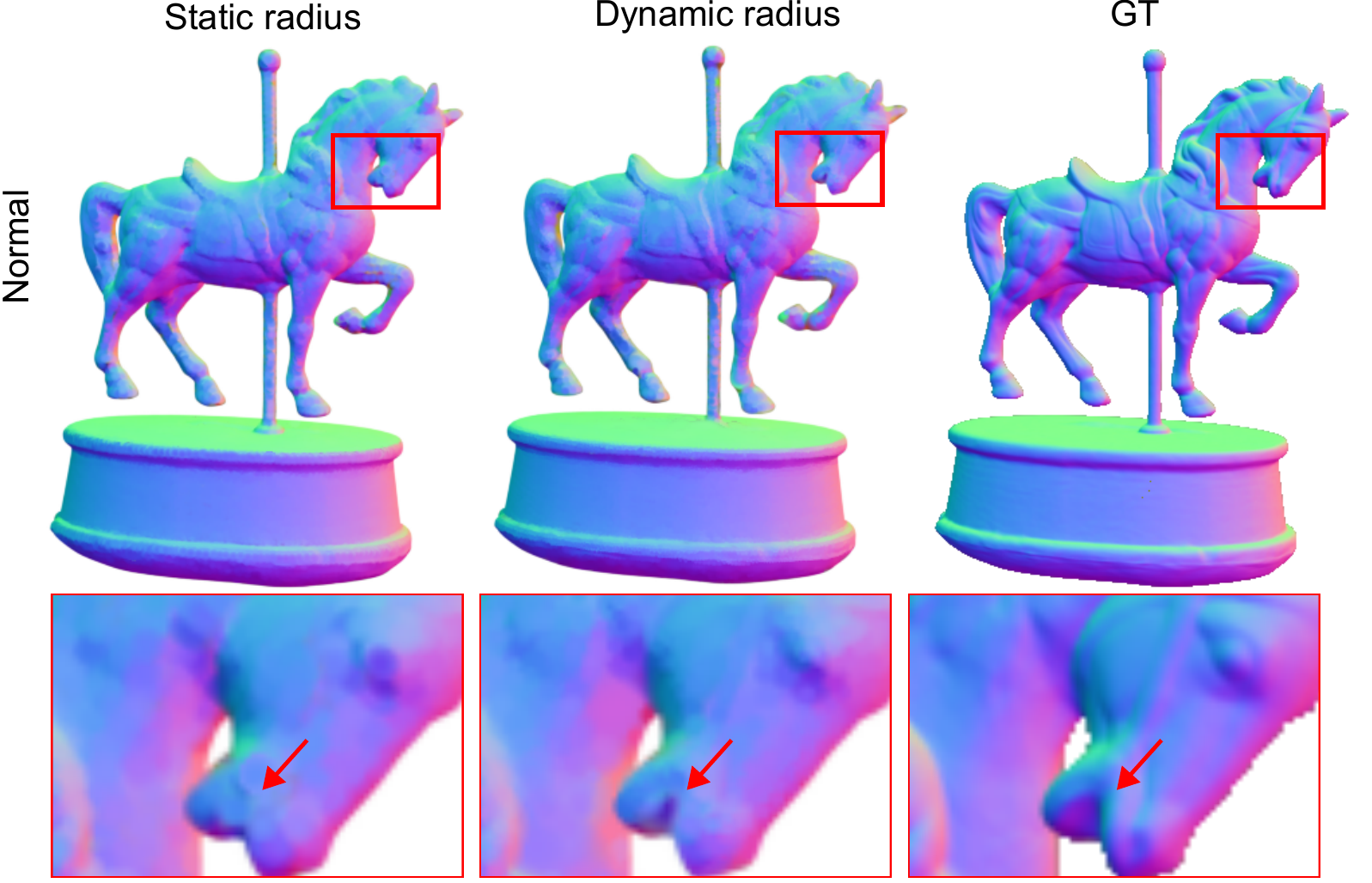}
    \caption{\textbf{Impact of point radius optimization.} Learning point radius allows for reconstructing geometry with both low- and high-frequency geometric features. 
    } 
\vspace{-2mm}
\label{fig:radius_parameter} 
\end{figure}

\paragraph{Dynamic Point Radius}
DPIR learns not only the location of points but also their radii, used for $\alpha$-blending in Equation~\eqref{eq:pixelcolor}.
Figure~\ref{fig:radius_parameter} show that learning the point radii enables accurate geometry reconstruction by adjusting the point size to the spatial frequencies of a target scene. 

\begin{figure}[t]
    \centering
    \includegraphics[width=\linewidth]{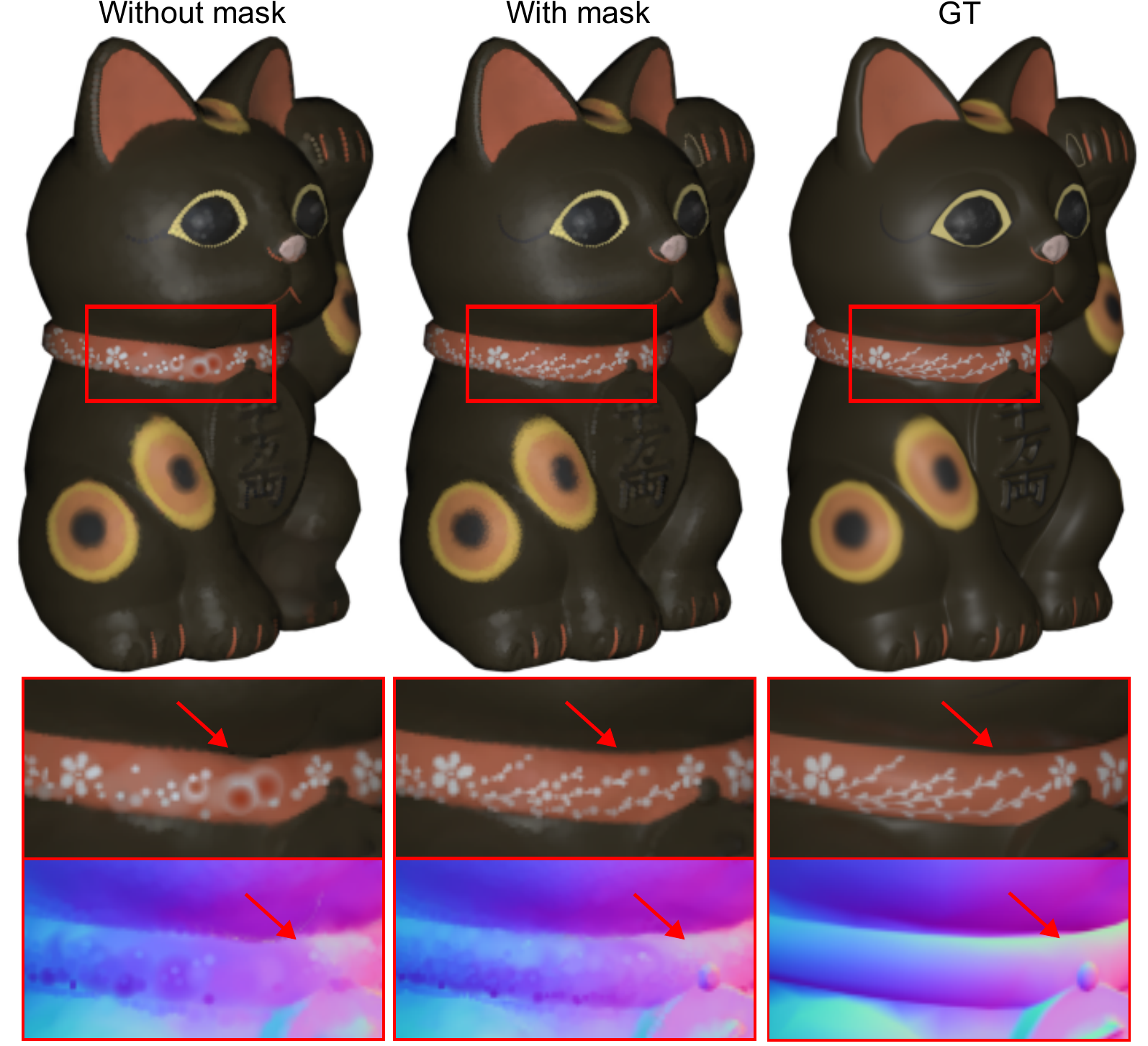}
    \caption{\textbf{Ablation on mask inputs.} Using mask inputs improves reconstruction quality while DPIR can obtain plausible results even without mask inputs. {Without using mask, optimization results near surface suffers from reconstruction artifacts. }
    }
\vspace{-4mm}
\label{fig:mask_init} 
\end{figure}

\paragraph{Dependency on Masks}
DPIR uses mask inputs to initialize the point locations and compute the mask loss. 
{Mask-based point initialization provides coarse point geometry as a starting step and the mask loss provides further shape guidance for low-light intensity region.}
In Figure~\ref{fig:mask_init}, we demonstrate that using mask inputs improves reconstruction quality. 
However, DPIR often achieves plausible reconstruction even without using the masks that show a potential application of DPIR for larger-scale scenes. 

\begin{figure}[t]
    \centering
    \includegraphics[width=\linewidth]{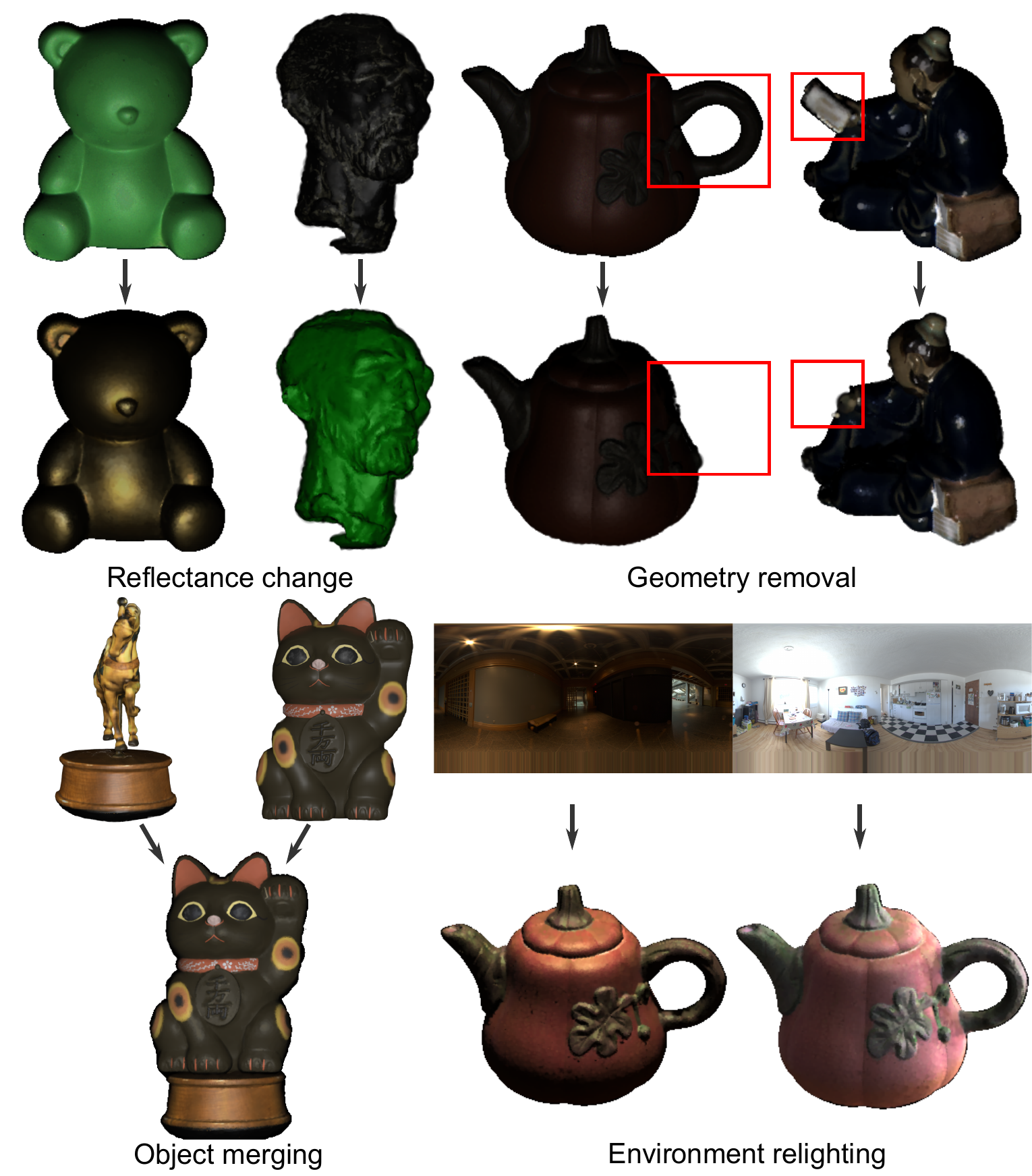}
    \caption{\textbf{Applications using point-based scene representation rendering.} DPIR enables convenient scene editing. We demonstrate reflectance change, geometry removal, object merging, and environment relighting.
    }
\vspace{-4mm}
\label{fig:applications} 
\end{figure}

\subsection{Applications}
The explicit point representation and rendering of DPIR enable convenient scene editing by simply editing points and its attributes.
Figure~\ref{fig:applications} shows four applications enabled by DPIR.
We swap the diffuse and specular basis BRDF of an object with that of another object we scanned. 
Also, simply editing the points from the reconstruction allows for intuitive geometric editing as we demonstrated in the geometry removal and the object merging.
Lastly, DPIR supports environment-map rendering by accumulating reflected radiance for each light source in the environment map. {Environment-map rendering time is linearly proportional to the number of illumination sources.}
\section{Conclusion}
\label{sec:Conclusion}

In this paper, we introduced DPIR, a point-based inverse rendering method that integrates efficient differentiable point-splatting forward rendering into the analysis-by-synthesis inverse rendering framework. To achieve this, we have developed a hybrid point-volumetric geometry representation, introduced regularized basis BRDFs, and used a point-based visibility detection method. DPIR jointly optimizes the point locations, radii, surface normals, and reflectance in a single stage without using any pre-trained network. Through evaluations, we demonstrate DPIR outperforms state-of-the-art inverse rendering methods in accuracy, training speed, and memory footprint.

\paragraph{Limitations}
Many interesting open questions remain. First, DPIR only models direct reflection without considering global-illumination effects such as inter-reflections. 
Modeling the global-illumination effects with an efficient global rendering method would be an interesting future work.
Second, DPIR does not explicitly model transmission. Generalizing inverse rendering with bidirectional scattering distribution functions may open up the applicability of DPIR for more diverse materials.

\paragraph{Acknowledgements}
This work was supported by Korea NRF (RS-2023-00211658, 2022R1A6A1A03052954), Samsung Advanced Institute of Technology, Samsung Research Funding \& Incubation Center for Future Technology grant (SRFC-IT1801-52), Samsung Electronics, and Korea IITP MSIT (RS-2023-00215700).

\clearpage
{\small
\bibliographystyle{ieee_fullname}
\bibliography{references}
}

\end{CJK}
\end{document}


\begin{CJK}{UTF8}{}
\CJKfamily{mj}

\title{Differentiable Point-based Inverse Rendering\\
\it Supplementary Information}

\author{
Hoon-Gyu Chung ~ ~ ~
Seokjun Choi ~ ~ ~
Seung-Hwan Baek \\
POSTECH\\
}
{
}

\maketitle
\hypersetup{linkcolor=black}
\noindent
This document offers extended information and further results of DPIR, complementing the main paper. 

\tableofcontents

\newpage

\section{Network Architecture}
\begin{figure*}[!ht]
    \centering
    \includegraphics[width=0.55\linewidth]{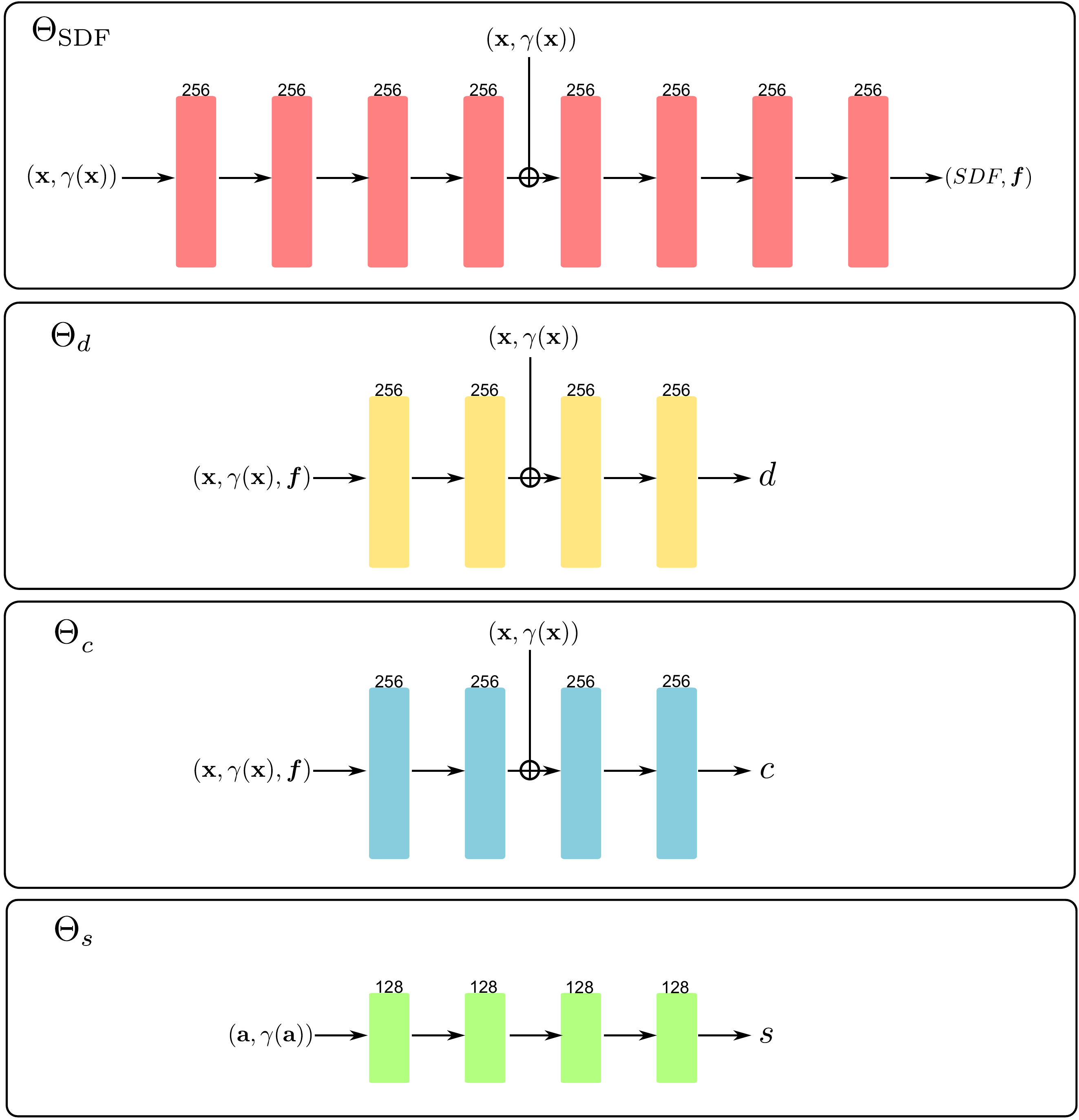}
    \caption{\textbf{Network architecture.} To compute point radiance, we estimate surface normal, diffuse albedo, and specularity using MLPs. Here, each block represents the fully-connected layer with its size of the hidden channel written at the top of the block. We use Softplus activation layer for $\Theta_\text{SDF}$ and ReLU activation layer for the other networks. $\oplus$ denotes skip connection that concatenates its output features with the input.}
\vspace{-5mm}
\label{fig:network} 
\end{figure*}

Figure~\ref{fig:network} shows the overall structure of of our DPIR network architecture which consists of four MLPs: (a) geometry network $\Theta_\text{SDF}$, (b) diffuse albedo network $\Theta_d$, (c) specular basis coefficient network $\Theta_c$, (d) specular basis network $\Theta_s$.

\subsection{Geometry Network} 
We use 8-layer MLPs of width 256 for SDF values and geometric features with skip connection at the 4th layer. We add positional encoding for input 3D point position $\mathbf{x}$ using 6 frequency components to train high frequency information. Positional encoding for point location $\mathbf{x}$ is represented as $\gamma (\mathbf{x})$.
\subsection{Diffuse Albedo Network}
We use 4-layer MLPs of width 256 for diffuse albedo with skip connection at the 2nd layer. We add positional encoding for input 3D point position $\mathbf{x}$ using 10 frequency components. Diffuse albedo network also utilizes SDF-based geometric features as input concatenated with positional-encoded point location.  
\subsection{Specular Basis Coefficient Network}
We use the same network architecture with diffuse albedo network. Specular basis coefficient network outputs 1 channel regularized values which are combinated with specular bases to compute specularity.
\subsection{Specular Basis Network}
We use 4-layer MLPs of width 128 for specular basis without skip connection. We compute SDF normal of each point and calculate half-way vector $\boldsymbol{h}$ considering viewing direction and light direction. We can represent isotropic BRDF with two parameters $(\theta_h,\theta_d)$ following ~\cite{efficient,shadow}, where $\theta_h = arccos(\mathbf{n}^{T}\mathbf{h})$, $\theta_d = arccos(\mathbf{v}^{T}\mathbf{h})$. We take the cosine value of these values and add positional encoding for two variables using 4 frequency components. Positional encoding for half-angle vectors $\mathbf{a}$ is represented as $\gamma (\mathbf{a})$.

\section{Loss Functions}
We optimize point positions $\mathbf{x}$, point radii $r$, and MLPs for SDF $\Theta_\text{SDF}$, diffuse albedo $\Theta_d$,  specular coefficient $\Theta_c$, and specular-basis $\Theta_s$ by minimizing the following loss function:
\begin{align}\label{eq:loss}
\mathcal{L}_2 + \lambda_\mathrm{ssim} \mathcal{L}_\mathrm{ssim}
+\lambda_{\mathrm{SDF}}  \mathcal{L}_\mathrm{SDF} + \lambda_{\mathrm{c}}   \mathcal{L}_\mathrm{c} +\lambda_{\mathrm{m}} \mathcal{L}_\mathrm{m}.
\end{align}

Here, $\mathcal{L}_2$ is the $\mathbf{l}_{2}$ loss for the reconstructed image $I^{'}$ and the observed image $I$. $\mathcal{L}_\mathrm{ssim}$ is the differentiable SSIM loss which considers luminance, contrast, structure for $I^{'}$ and $I$. Our fast splatting-based forward rendering allows DPIR to utilize SSIM loss, which is often omitted in other rendering methods due to its long computation time, resulting better reconstruction quality. $\mathcal{L}_\mathrm{SDF}$ is to promote the zero-level set of SDF exists near explicit point positions, combining discrete point representation with continuous SDF. $\mathcal{L}_\mathrm{c}$ regularizes $\mathbf{l}_{1}$ norm of estimated per-point specular coefficients to be $\epsilon$. Specular coefficients are constrained to be positive and under $\epsilon$. $\mathcal{L}_\mathrm{m}$ is the $\mathbf{l}_{2}$ loss for the reconstructed mask and the ground truth mask. Reconstructed mask image is rendered by point-based splatting with radiance of 1 for all points. We set $\lambda_\mathrm{ssim}$, $\lambda_{\mathrm{SDF}}$, $\lambda_{\mathrm{c}}$ and $\lambda_{\mathrm{m}}$ as 0.2, 1.0, 0.1 and 0.1, respectively.

\section{Optimization Techniques}
\subsection{Mask-based Point Initialization}
For point cloud initialization, we employ rejection sampling method which samples 3D points uniformly, project these points on each image plane, and let them fall within all the masks~\cite{differentiable}. Mask-based point initialization provides coarse geometry for efficient and stable optimization, thus DPIR reconstruction quality is dependent on the accuracy of the mask inputs. We show ablation study that mask inputs improves reconstruction quality while DPIR can obtain plausible reconstruction result without mask.

\subsection{Coarse-to-fine Updates}
Our method adopts coarse-to-fine updates to learn accurate point cloud representations for geometry and reflectance. First, we employ a voxel discretization to combine points within same voxel into one single point. Second, we compute the distances between aggregated points and standard deviation of these distances. We then remove outliers whose standard deviation is beyond threshold. Voxel-based downsampling enables pruning of superfluous points. After we prune unnecessary points, we insert new points into the point cloud by upsampling remained point representations with same parameters. We repeat these stages for 5 times to achieve coarse-to-fine updates with stable and fast training. Both of mask-based initialization and coarse-to-fine updates are inspired by ~\cite{differentiable}.

\subsection{Training}
To train our DPIR method, we empirically choose sampling rate of the number of initialized points and the initialized point radius considering the size of each object. We train 40 epochs for every stage and do not consider visibility at the first stage. Our DPIR method is trained for 7 stages which take 2 hours to converge. We use Adam as optimizer and set the initial learning rate for the point parameters and network parameters as 1e-4 and 5e-4, respectively. Both of learning rates are decayed exponentially for every epoch with a factor of 0.93. We use PyTorch and test DPIR on a single NVIDIA RTX 3090 GPU.
\section{Dataset}
\subsection{DiLiGenT-MV}
We test our DPIR method on DiLiGenT-MV, multi-view multi-light image dataset which is often used for evaluating multi-view photometric stereo. DiLiGenT-MV dataset consists of 5 objects, called Bear, Buddha, Cow, Pot2, Reading. For each object, images are captured under horizontally rotated 20 cameras with same angle and same elevation. For each view, 96 images are captured under different single directional light source with different light intensity. We preprocessed images to make low intensity images brighter by normalizing images according to the ground truth light intensity. We also cropped the original images with 612 $\times$ 512 into 400 $\times$ 400 to remove empty background region for efficiency. Our pre-processed dataset followed the instructions from PS-NeRF~\cite{psnerf}.

\subsection{Synthetic Photometric Dataset}
We also test our DPIR method on synthetic photometric dataset, following the configuration of mobile flash photography ~\cite{iron}. We rendered 4 objects, called Dragon, Head, Horse, Maneki, with Blender using mesh and image texture data from IRON~\cite{iron}. We rendered 300 views images with co-located point lights and used 200/100 views for training/testing, respectively. It took around 3 hours to render ground truth images and normal for 300 views.

\section{Additional Ablation Study}
\begin{table*}[h]
\centering
\resizebox{0.8\linewidth}{!}{
    \begin{tabular}{c|cccc|cccc}
    \toprule[1pt]
            &\multicolumn{4}{c|}{Multi-view multi-light dataset} & \multicolumn{4}{c}{Photometric dataset} \\ \hline
    Method &PSNR $\uparrow$&SSIM $\uparrow$&LPIPS $\downarrow$   &MAE $\downarrow$        &PSNR $\uparrow$&SSIM $\uparrow$&LPIPS $\downarrow$   &MAE $\downarrow$     \\
    \hline
    Proposed   & \textbf{36.73} & \textbf{0.9822} & \textbf{0.0091} & 7.16 & \textbf{35.56} & 0.9734  & 0.0285 &  \textbf{8.74} \\ \hline
    w/o vis    & 35.17 & 0.9756 & 0.0126 & 9.57 & x & x  & x &  x \\
    w/o SDF   & 33.39 & 0.9704 & 0.0156  & 21.89 & 34.55 & 0.9655  & 0.0378 & 19.42   \\ 
    w/o radii  & 26.62 & 0.9616 & 0.0392 & 10.48   & 35.48   & \textbf{0.9740}  & 0.0295 & 8.77  \\
    w/o $\mathcal{L}_\mathrm{ssim}$ &32.97  &0.9738  &0.0132 & 9.01   & 34.75 &0.9718 &0.0373  & 10.32   \\
    basis 1  & 36.12 & 0.9816 & 0.0096 & 7.41 & 35.12 & 0.9712  & 0.0303 &  9.31    \\ 
    basis 5   & 36.44 & 0.9820 & 0.0094 & 7.22 & 35.45 & 0.9731  & 0.0287 &  8.76  \\
    basis 13   & 36.40 & 0.9821 & 0.0092 & \textbf{7.11} & 35.41 & 0.9732  & \textbf{0.0283} &  8.77\\ 
    \bottomrule[1pt]
    \end{tabular}}
    \caption{Quantitative comparison of ablation studies for multi-view multi-light dataset and photometric dataset.}
\vspace{-5mm}
\label{tab:ablation_table}
\end{table*}
\subsection{Point-based Shadow Detection}
We evaluate the importance of the point-based visibility test of DPIR. Note that visibility of every point is set to 1 on photometric dataset as the light source and the camera are co-located. Table~\ref{tab:ablation_table} shows that point-based shadow detection method improves both image and normal reconstruction quality. Especially, normal estimation of self-occluded region is enhanced.

\subsection{Hybrid Shape Representation}
We evaluate the impact of hybrid point-volumetric shape representation. Table~\ref{tab:ablation_table} shows that using only point representation and per point normal for inverse rendering recurs inaccurate reconstruction. Our hybrid point-volumetric representation improves normal reconstruction quality by sampling surface normals of discrete points from continuous SDF

\subsection{Dynamic Point Radius}
We evaluate the impact of point radius optimization. Table~\ref{tab:ablation_table} shows that learning not only the point position but also point radius enables accurate geometry reconstruction for low and high frequency details. Dynamic point radius shows reconstruction improvement of large margin especially on multi-view multi-light dataset.

\subsection{Number of Basis}
\begin{figure*}[h]
    \centering
    \includegraphics[width=0.65\linewidth]{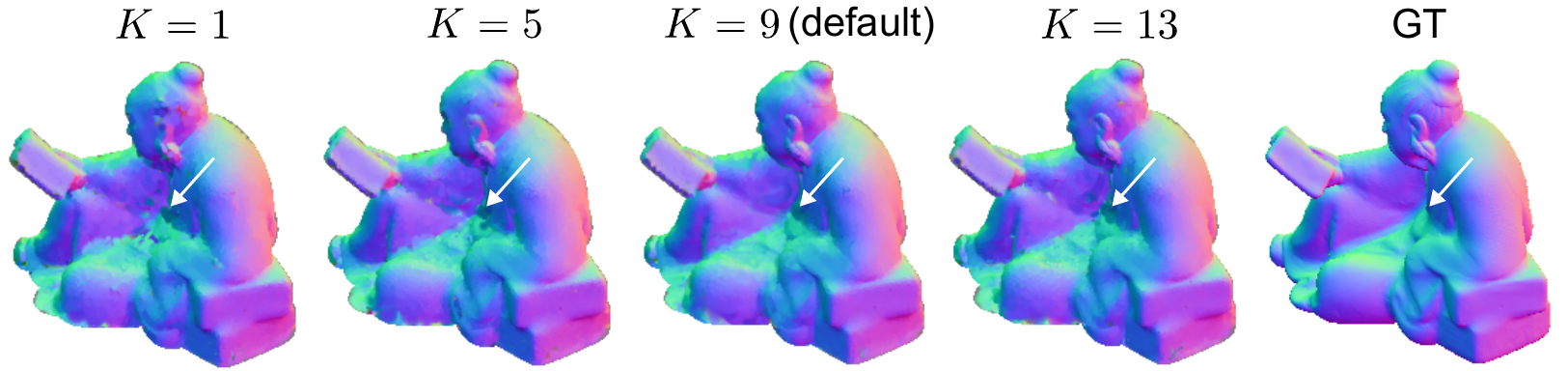}
    \caption{\textbf{Impact of the number of specular basis.} Increasing the number of specular basis improves reconstruction quality and here using nine bases shows converged performance.}
\vspace{-5mm}
\label{fig:number_basis} 
\end{figure*}

We evaluate the impact of the number of specular basis. Table~\ref{tab:ablation_table} and Figure~\ref{fig:number_basis} show that using 9 bases provides a converged accuracy in a tested scene. Using few specular basis such as one and five results in inaccurate reconstructions. The number of specular basis is related to the representation power of specularity and normal. 
\newpage
\subsection{Impact of SSIM Loss}
We evaluate the importance of SSIM loss which is computationally expensive. Our DPIR method adopts SSIM loss for better reconstruction based on fast splatting-based rendering. It requires 0.15$\times$ additional training time, while improving image and normal reconstruction quality for a large margin.  

\subsection{Number of Training Views and Lights}
Table~\ref{tab:view_light} shows the normal reconstruction accuracy with varying number of training views and lightings. 
We found that the number of views plays an important role while the light gives a smaller impact when the number of training lights exceeds 16. Our method can achieve state-of-the-art reconstruction result when trained with only 16 lights.
\begin{table}[h]
\centering
\resizebox{0.6\columnwidth}{!}{
   \begin{tabular}{l|cccc}
    \toprule[1pt]
    
                        &{4 Lightings} & {10 Lightings} & {16 Lightings} &{30 Lightings}  \\
    \hline
    5 Views             &22.78    &18.89    &17.02    &15.53            \\
    10 Views            &19.04    &13.69    &11.53    &10.05              \\
    15 Views            &14.63    &10.15    &8.88     &8.22            \\ 
     \bottomrule[1pt]
    \end{tabular}}
    \caption{Impact of the number of views and lightings for "Reading", in terms of normal reconstruction with MAE.} 
\vspace{-5mm}
\label{tab:view_light}
\end{table}

\subsection{Threshold $\tau$ for visibility test}
{We assumed that points are located inside a unit cube according to the given camera parameters of real-world dataset: $\tau$ is defined in a normalized scale. 
Table~\ref{tab:threshold_comparison} shows that $\tau=0.1$ leads to overall lower normal reconstruction error in MAE than $\tau=0.2$ and $\tau=0.05$. Hence, we chose $\tau=0.1$. 
Our joint optimization provides accurate reconstruction of depth, reflectance, and normal, resulting in accurate visibility. }
\begin{table}[h]
\centering
\vspace{-3mm}
\resizebox{0.5\linewidth}{!}{
    \begin{tabular}{c|ccccc}
    \toprule[1pt]
           &\multicolumn{1}{c}{Bear} & \multicolumn{1}{c}{Buddha} & \multicolumn{1}{c}{Cow} & \multicolumn{1}{c}{Pot2} & \multicolumn{1}{c}{Reading}  \\
    Scene  & {MAE $\downarrow$}   & {MAE $\downarrow$}    & {MAE $\downarrow$}    & {MAE $\downarrow$}   & {MAE $\downarrow$}   \\ 
    \hline
    $\tau=0.2$                   & 4.62 & 11.37 & \textbf{4.48} & 6.63 & 9.21  \\
    $\tau=0.05$                 & 6.30 & 14.78 & 4.66 & \textbf{6.54} & 9.37  \\ 
    \hline
    $\tau=0.1$                   & \textbf{4.35} & \textbf{11.10} & 4.61 & 6.71 & \textbf{9.03}     \\ 
    \bottomrule[1pt]
    \end{tabular}}
    \vspace{-2mm}
    \caption{Ablation study for point-based visibility test threshold.}
\vspace{-6mm}
\label{tab:threshold_comparison}
\end{table}

\subsection{Mask Dependency}
\begin{figure*}[h]
    \centering
    \includegraphics[width=0.45\linewidth]{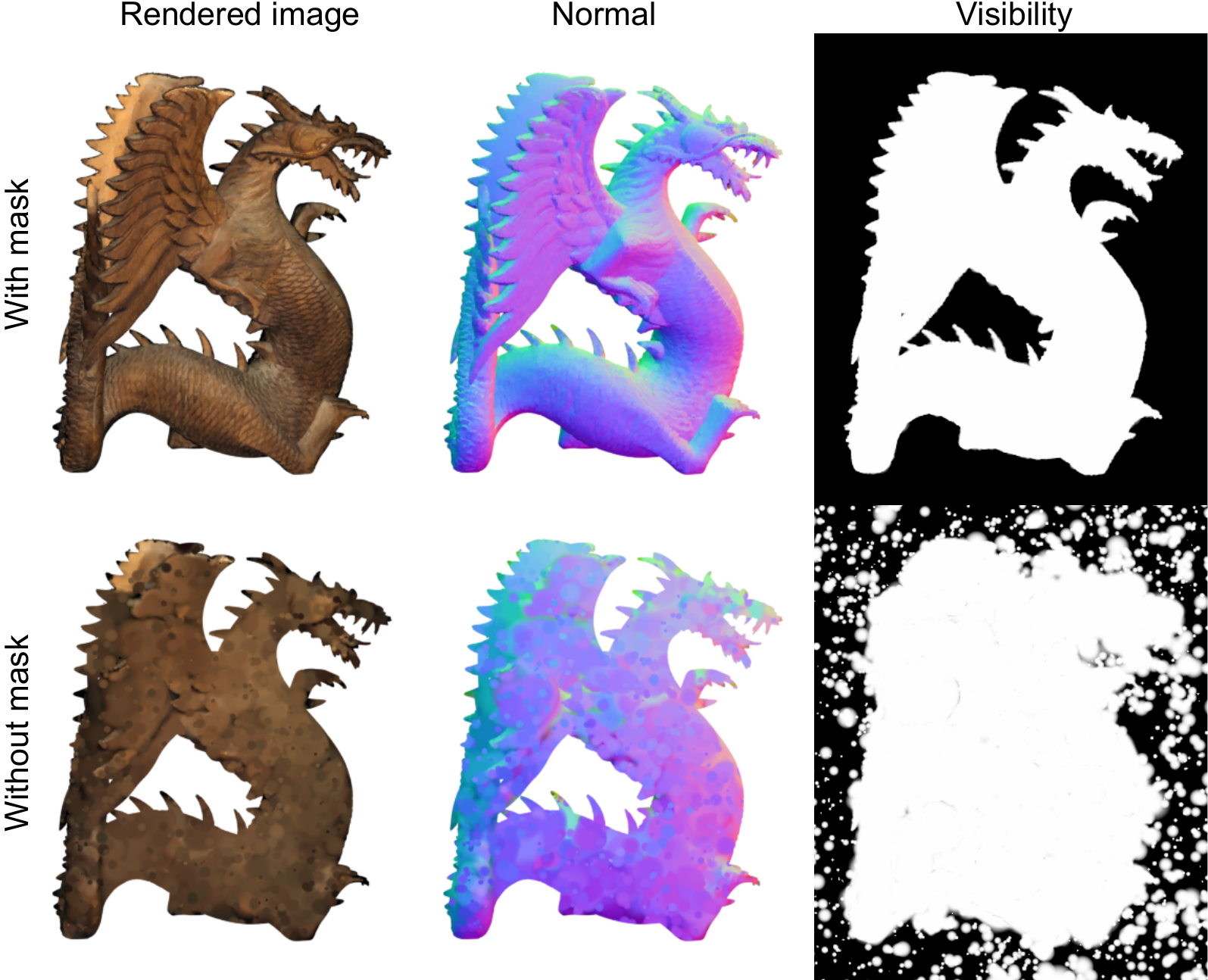}
    \caption{\textbf{Importance of mask inputs.} We test our DPIR without mask inputs, initializing point positions with randomly sampled points inside unit sphere. For complex scene as "Dragon", point positions are noisy and diverge.
    }
\vspace{-5mm}
\label{fig:supple_ablation_mask} 
\end{figure*}

DPIR utilizes mask inputs for point location initialization and mask loss. Our method often achieves plausible reconstruction results even without masks inputs which show the potential applicability of DPIR for larger-scale scene. For complex geometry object, our point locations diverge as shown in Figure~\ref{fig:supple_ablation_mask}, meaning our optimization techniques are incomplete. Developing our optimization techniques for complex scene without mask inputs is our future work.

\section{Additional Discussions}
\subsection{Specular Basis BRDFs and Specular Coefficients}
\begin{figure*}[h]
    \centering
    \includegraphics[width=0.7\linewidth]{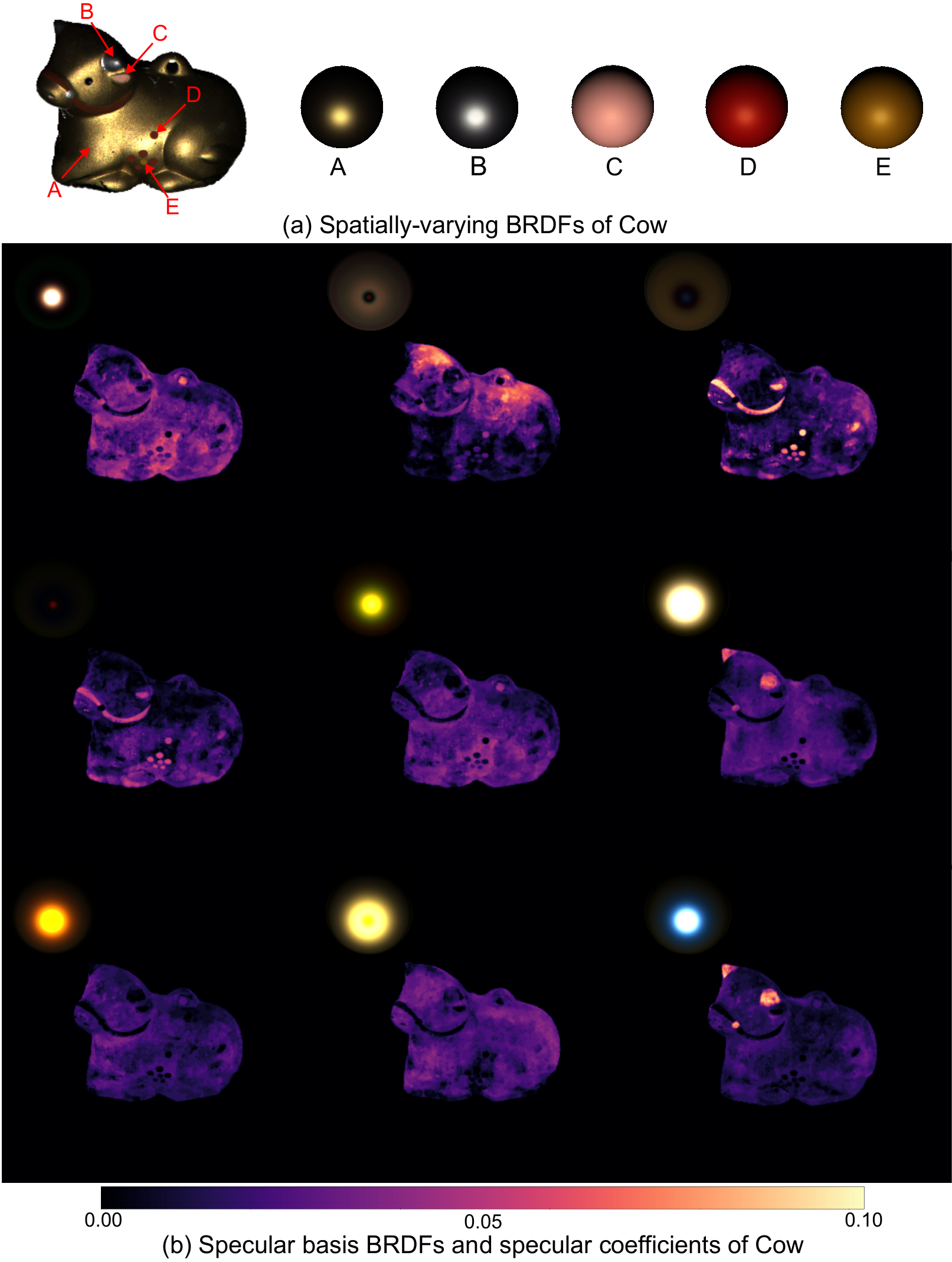}
    \caption{\textbf{Estimated spatially-varying BRDFs, specular basis BRDFs, and specular coefficients.} We visualize the BRDFs on unit spheres illuminated by a point light source. (a) shows spatially-varying BRDFs of "Cow" with diverse appearance. (b) shows specular basis BRDFs and corresponding specular coefficients.
    }
\vspace{-5mm}
\label{fig:basis_coeff1} 
\end{figure*}

We utilize regularized basis BRDF representation to estimate accurate spatially-varying BRDFs from limited light-view angular samples. Figure.~\ref{fig:basis_coeff1} shows visualization of spatially-varying BRDFs, specular basis BRDFs, and specular coefficients of "Cow". Specular basis BRDFs for gold appearance have high specular coefficients for body region. Specular basis BRDFs for silver appearance have high specular coefficients for horn region. Specular coefficients for diffuse dominant region as red and yellow have low intensity of specular basis BRDFs. 

\subsection{Evaluation Metrics with Mask}
For fair quantitative comparison between different baselines, mask computation performs critically especially on the mean angular error (MAE). We used rendered mask region for calculating the MAE, while mask estimation of each baseline is different with ground-truth mask. We calculate the MAE with rendered normal and ground-truth normal using rendered mask. Image reconstruction metrics (PSNR, SSIM, LPIPS) are calculated with white background images.

\subsection{Specularity with Shading}
\begin{figure*}[h]
    \centering
    \includegraphics[width=0.38\linewidth]{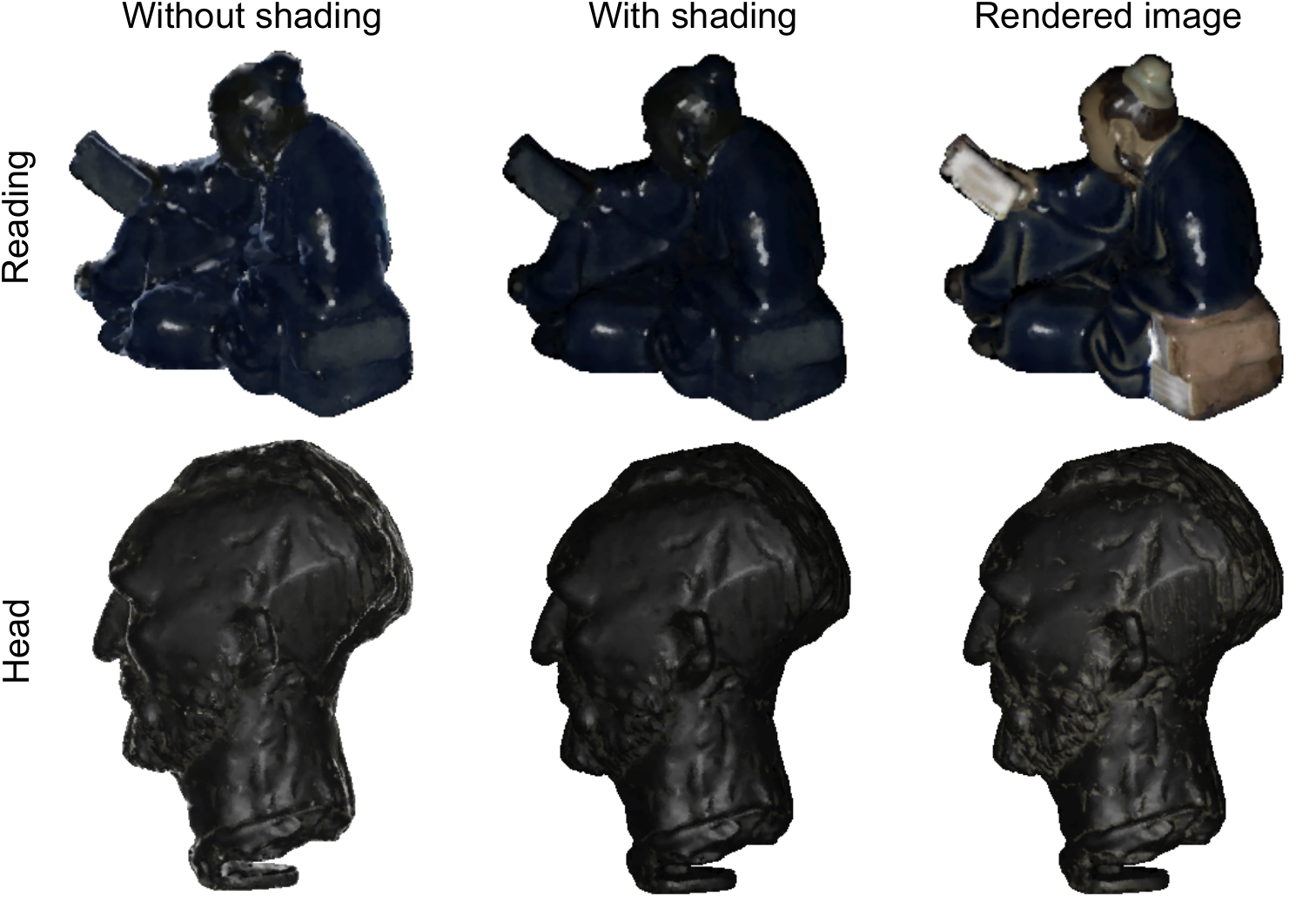}
    \caption{\textbf{Specularity images with and without shading.} Specularity images without shading have more artifacts at the edge of the object, while specularities are more stable with shading.
    }
\vspace{-6mm}
\label{fig:specularity} 
\end{figure*}

We visualize specularity image with shading which denotes cosine value between normal and light direction. Our DPIR method computes the point radiance with point BRDF and shading. 
Thus, we render specularity image by computing point radiance consisting of specular BRDF and shading. Figure.~\ref{fig:specularity} shows ablation study with shading.

\section{Additional Results}
\subsection{Gaussian-based Inverse Rendering}
We compare DPIR with concurrent work: Gaussian-based inverse rendering method, Rel-GS ~\cite{gao2023relightable}.
Rel-GS takes images captured under a constant environment map, similar to PhySG and TensoIR. 
To meet its input requirements, we use multi-light averaged images and compare Rel-GS with DPIR. 
Table~\ref{tab:reb_table} shows that DPIR outperforms Rel-GS in  rendering quality and normal accuracy. 
\begin{table}[h]
\centering
\vspace{-4mm}
\resizebox{0.4\linewidth}{!}{
    \begin{tabular}{c|ccc|c}
    \toprule[1pt]
            &\multicolumn{4}{c}{Multi-view multi-light dataset} \\ \hline
    Method &PSNR $\uparrow$&SSIM $\uparrow$&LPIPS $\downarrow$   &MAE $\downarrow$ \\
    \hline
    Rel-GS    & 31.18 & 0.9687 & 0.0254 & 21.54 \\ \hline 
    DPIR   & \textbf{38.83} & \textbf{0.9908} & \textbf{0.0038} & \textbf{7.16} \\
    \bottomrule[1pt]
    \end{tabular}}
    \caption{Comparison between DPIR and Rel-GS}
\vspace{-4mm}
\label{tab:reb_table}
\end{table}

\subsection{Additional Visualization of Normals and Albedo for Figure~8}
\begin{figure}[h]
\vspace{-3mm}
    \centering
    \includegraphics[width=0.6\linewidth]{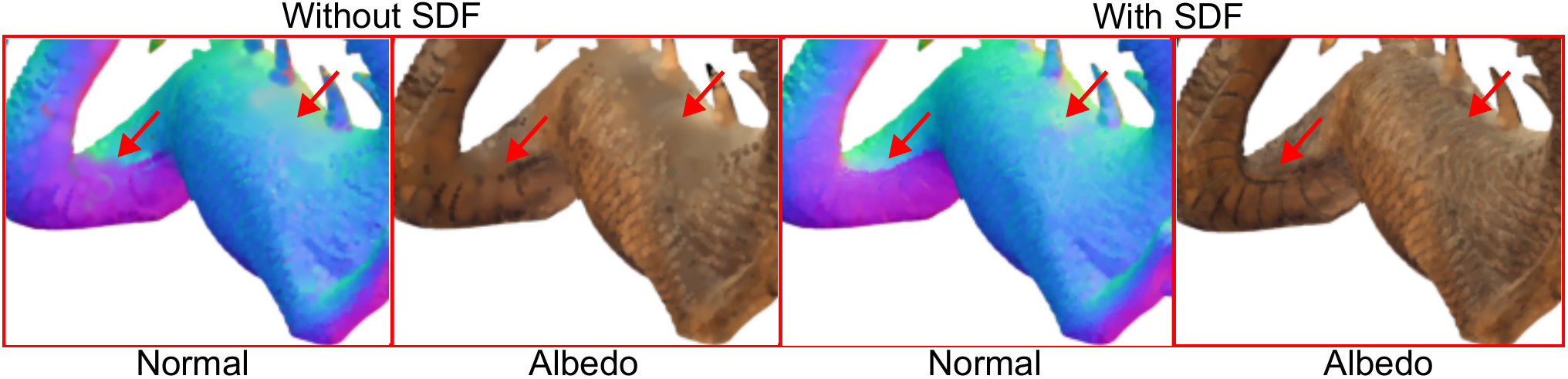}
    \caption{{Impact of hybrid shape representation.}}
\vspace{-3mm}
\label{fig:ablation_SDF_rebuttal} 
\end{figure}

Figure~8 in the main paper shows the rendered images, not the albedo.  
Figure~\ref{fig:ablation_SDF_rebuttal} shows both normals and albedo of the same scene, where details can be seen at both normals and albedo. 

\subsection{Environment Map Relighting}
\begin{figure*}[h]
    \centering
    \includegraphics[width=0.7\linewidth]{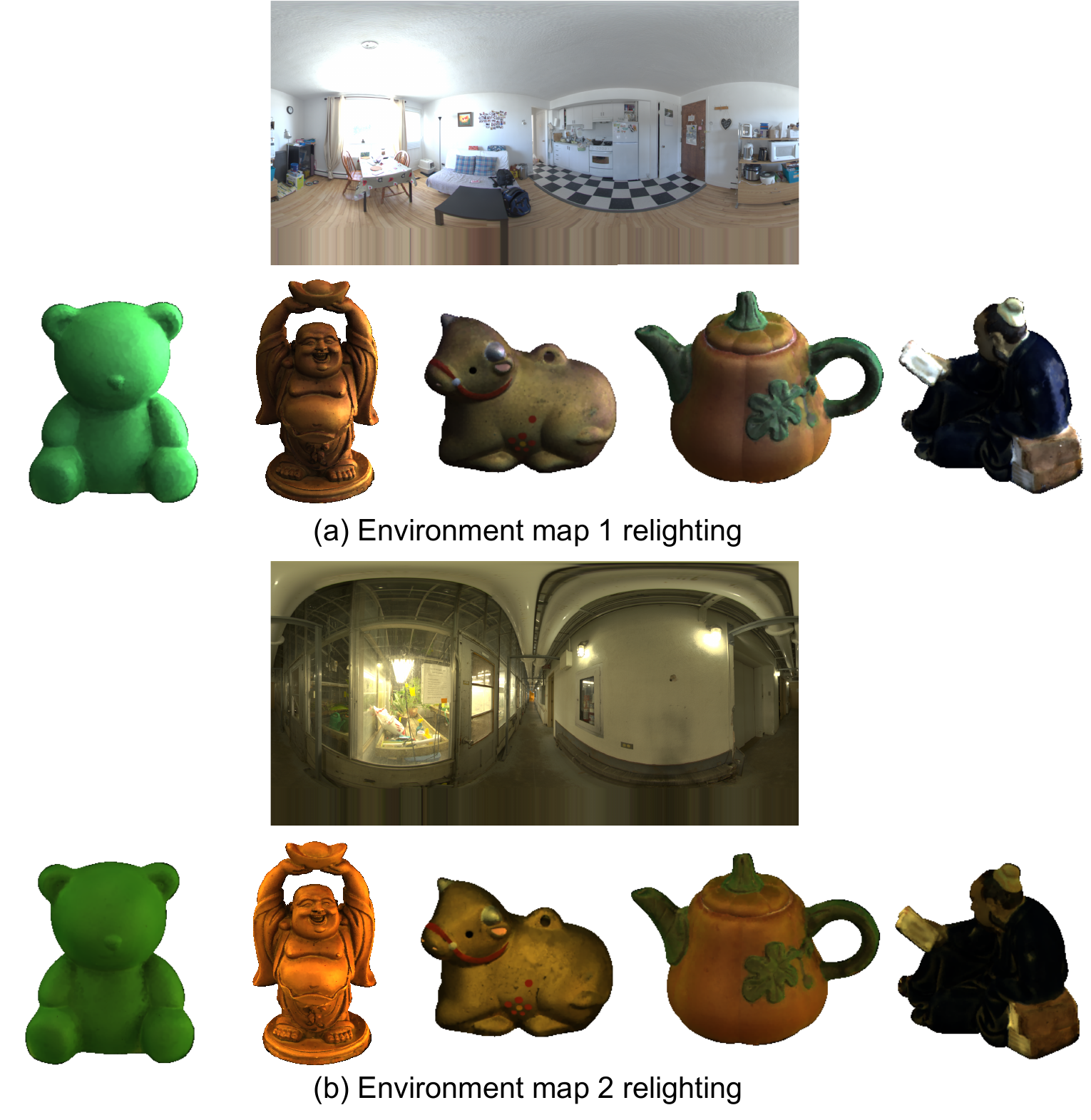}
    \caption{\textbf{Environment map rendering.} We render various objects with two environment maps. Both of rendering results show faithful relighting by reconstructed surface normals, spatially varying BRDFs, and visibility.
    }
\vspace{-2mm}
\label{fig:envmap_relighting} 
\end{figure*}

Our DPIR method allows environment map relighting by integrating reflected radiance for each light source in the environment map. In Fig.~\ref{fig:envmap_relighting}, "Bear", "Buddha", "Cow", "Pot2" and "Reading" are rendered with different environment map. They show faithful relighting results based on the environment map.
\subsection{Additional Results with Multi-view Multi-light Dataset}
\begin{figure*}[h]
    \centering
    \includegraphics[width=0.8\linewidth]{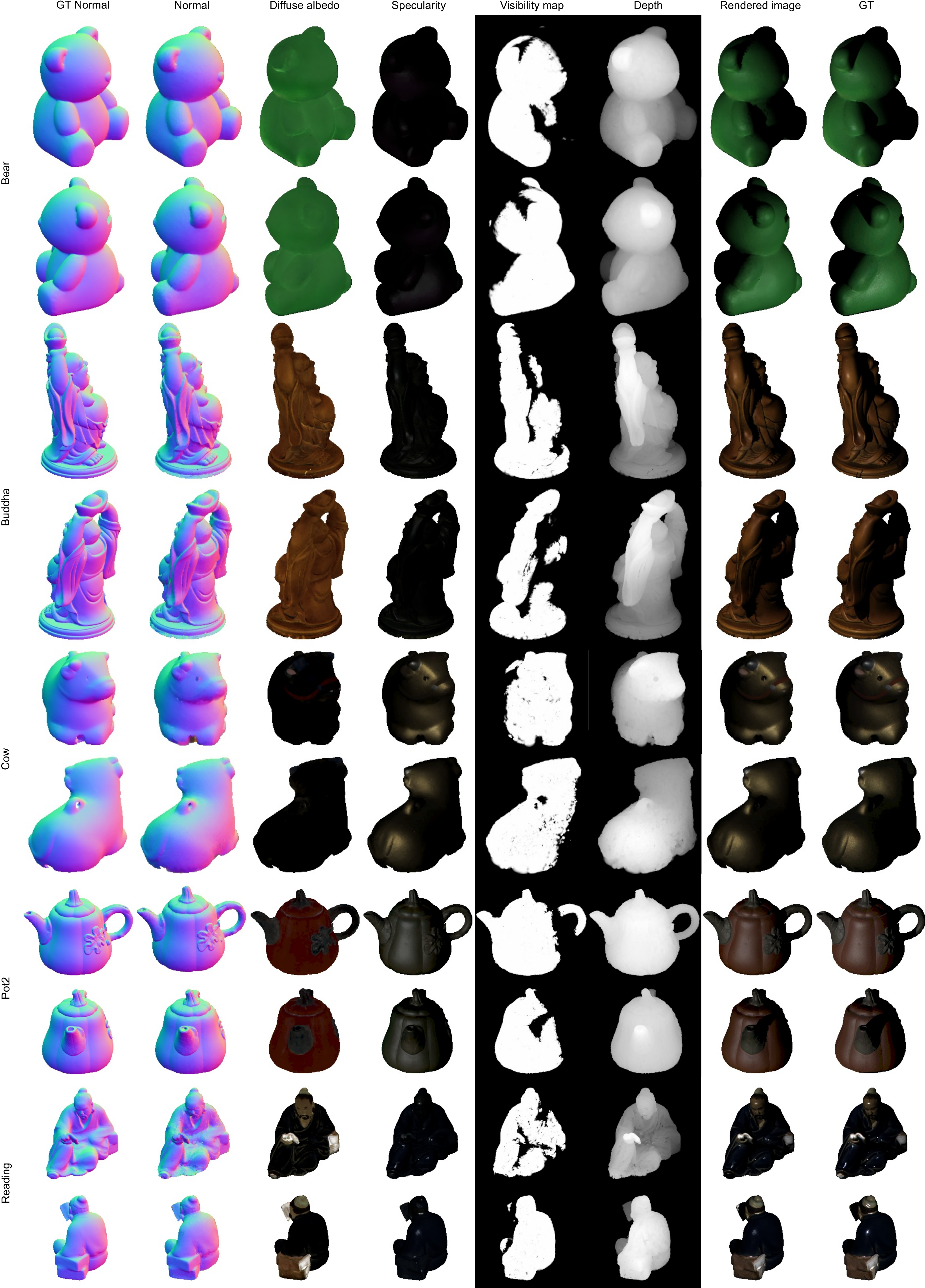}
    \caption{\textbf{Reconstruction results of DiLiGenT-MV objects.} We visualize rendered image, normal, diffuse albedo, specularity, visibility map, depth map with ground-truth normal and image. We perform novel-view relighting with various view points and lightings. In "Bear", the object has smooth surface, and visibility maps are well estimated with point-based shadow detection. "Buddha" has complex geometry and produces self-occlusion at head region. In "Cow", the object consists of metal and has difficulty in reconstructing eye region. In "Pot2", high frequency details are well reconstructed such as top and leaf of the object. "Reading" has concave geometry and produces high specular effects. Our DPIR method shows high quality surface and material reconstruction. We scale "Cow" and "Pot2" images brighter using gamma} 
\vspace{-10mm}
\label{fig:supple_diligent1} 
\end{figure*}

\begin{figure*}[t]
    \centering
    \includegraphics[width=\linewidth]{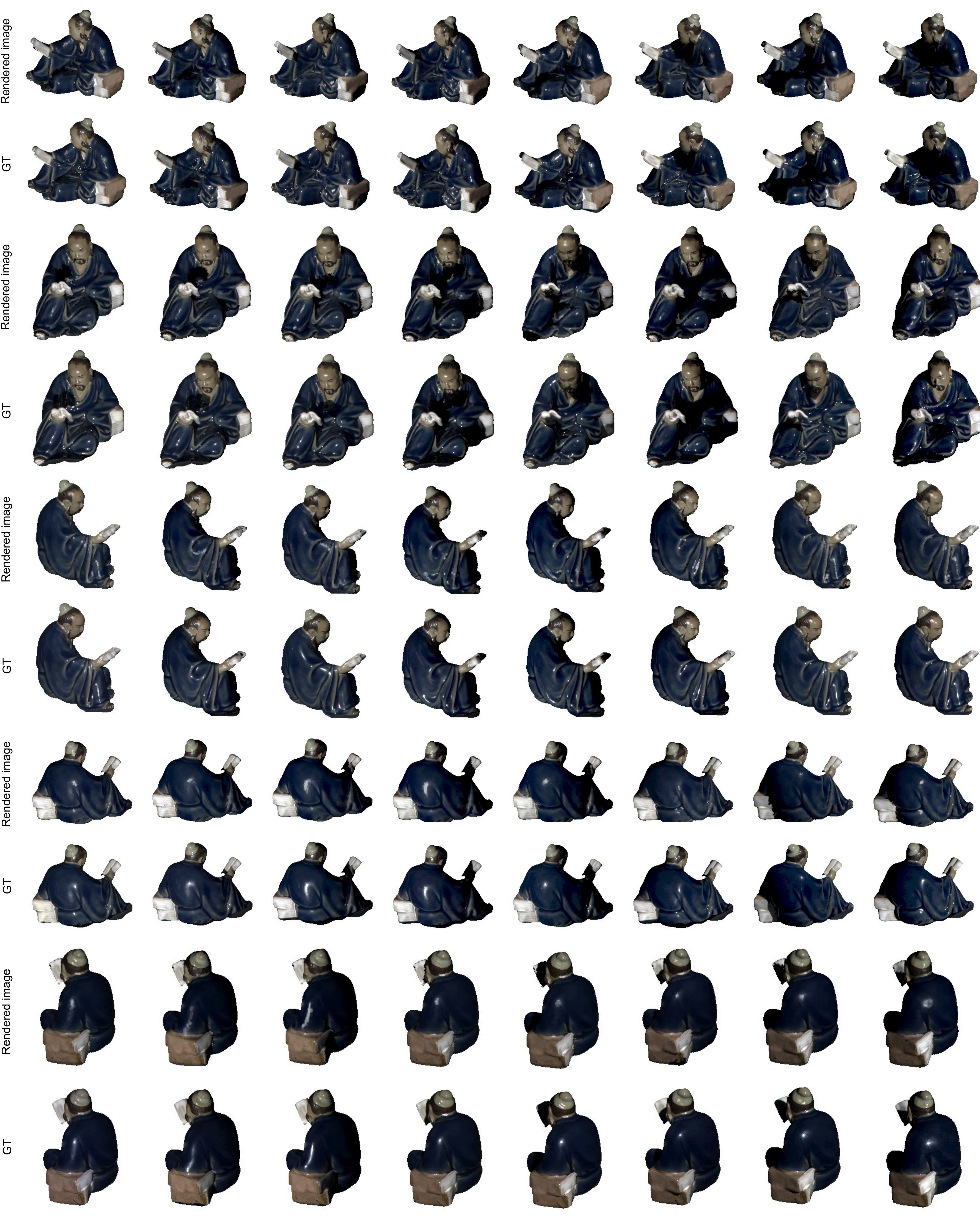}
    \caption{\textbf{Novel view relighting.} We visualize rendered images of 5 novel view points and 8 novel light sources. Our DPIR method achieves faithful reconstruction of shape and material for untrained scene. We scale up reconstructed images and ground-truth images for visualization.} 
\label{fig:novelview_relighting} 
\end{figure*}

Figure~\ref{fig:supple_diligent1} shows visualization of all objects from DiLiGenT-MV dataset. Our DPIR method achieves the best shape and material reconstruction results with multi-view multi-light dataset. Hence, we provide the visualization of the rendered image, estimated normal, diffuse albedo, specularity, visibility, and depth map. It demonstrates that our method is robust to diverse shapes and materials in real-world objects. Figure~\ref{fig:novelview_relighting} shows novel view relighting of 5 view points and 8 light directions.
\clearpage
\subsection{Additional Results with Photometric Dataset}
\begin{figure*}[h]
    \centering
    \includegraphics[width=0.8\linewidth]{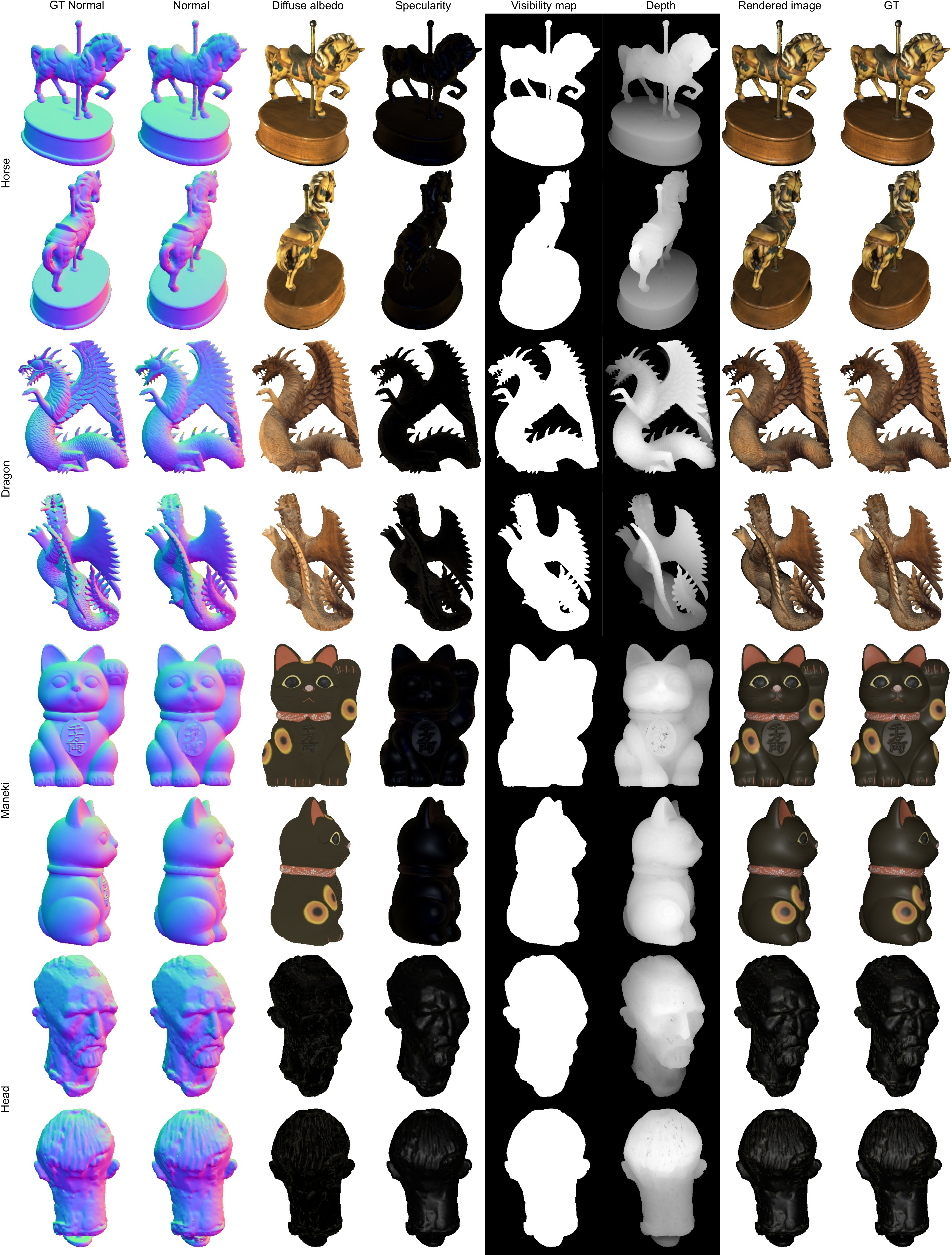}
    \caption{\textbf{Reconstruction results of photometric dataset.} We perform novel-view relighting on various view points with co-located point lights.} 
\label{fig:supple_photometric1} 
\end{figure*}

Figure~\ref{fig:supple_photometric1} shows visualization of all objects from synthetic photometric dataset. Our DPIR method achieves the state-of-the-art reconstruction results on different views with co-located point lights. Both of evaluations with different datasets demonstrate that our method is applicable to diverse illumination settings with flexible number of view points. Small number of view points can be compensated by the number of illuminations.
{\small
\bibliographystyle{ieee_fullname}
\bibliography{references}
}

\end{CJK}